\documentclass[11pt]{article}

\usepackage[preprint]{acl}

\usepackage{times}
\usepackage{latexsym}

\usepackage[T1]{fontenc}

\usepackage[utf8]{inputenc}

\usepackage{microtype}

\usepackage{inconsolata}

\usepackage{graphicx}

\usepackage{booktabs,siunitx,threeparttable,subcaption}
\usepackage[table,dvipsnames]{xcolor} 
\usepackage{multirow}

\title{Just Use XML: Revisiting Joint Translation and Label Projection}

\author{Thennal D K \and Chris Biemann \and Hans Ole Hatzel \\
  Language Technology Group \\
  University of Hamburg \\
  \texttt{thennal10@gmail.com} \\ 
  \texttt{\{chris.biemann, hans.ole.hatzel\}@uni-hamburg.de}}

\begin{document}
\maketitle
\begin{abstract}
Label projection is an effective technique for cross-lingual transfer, extending span-annotated datasets from a high-resource language to low-resource ones. Most approaches perform label projection as a separate step after machine translation, and prior work that combines the two reports degraded translation quality. We re-evaluate this claim with LabelPigeon, a novel framework that jointly performs translation and label projection via XML tags. We design a direct evaluation scheme for label projection, and find that LabelPigeon outperforms baselines and actively improves translation quality in 11 languages. We further assess translation quality across 203 languages and varying annotation complexity, finding consistent improvement attributed to additional fine-tuning. Finally, across 27 languages and three downstream tasks, we report substantial gains in cross-lingual transfer over comparable work, up to +40.2 F1 on NER. Overall, our results demonstrate that XML-tagged label projection provides effective and efficient label transfer without compromising translation quality.\footnote{Our code and data are available at: \url{https://github.com/thennal10/LabelPigeon}}
\end{abstract}

\section{Introduction}

\begin{figure*}[t]  \includegraphics[width=\textwidth]{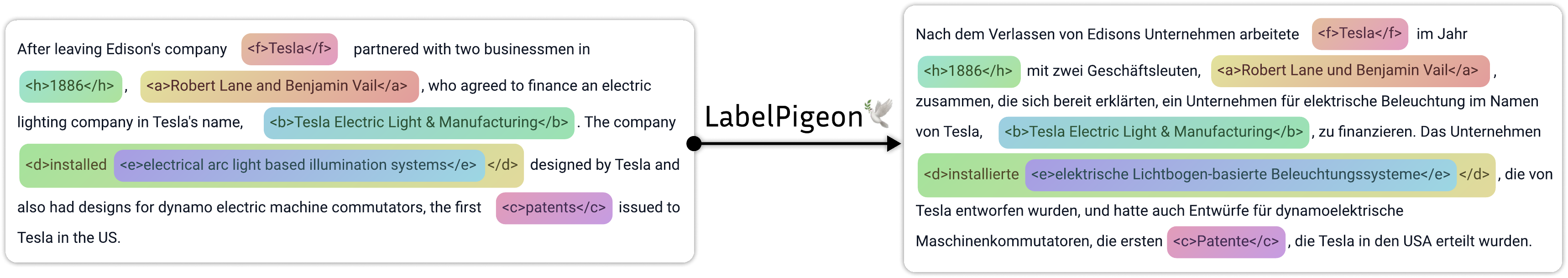}
  \caption{An example taken from XQuAD \citep{artetxeCrosslingualTransferabilityMonolingual2020}, where LabelPigeon accurately and seamlessly handles translating English to German while transferring 7 labeled spans with nesting.}
  \label{fig:labelpigeon}
\end{figure*}

Many NLP tasks depend on span-level labels, such as entities in named entity recognition, arguments in event extraction, or mentions in coreference resolution \citep{liuStructuredSpanSelector2022}. 
Although recent advances in generative large language models showcase strong zero-shot potential, supervised training on task-specific data continues to achieve substantially superior performance in a multilingual setting \citep{weiAreLLMsGood2024, poradaControlledReevaluationCoreference2024a, luLargeLanguageModels2025, bucherFineTunedSmallLLMs2024}. A common paradigm for extending these tasks beyond high-resource languages like English is the use of automatic machine translation to translate training data into the target language. This involves \textit{label projection}, techniques to preserve or subsequently map the span labels onto the translated text  \citep{chenFrustratinglyEasyLabel2024, ebingDevilWordAlignment2025a}.

Label projection has been traditionally conducted as a separate step from translation, largely with the use of word alignment models \citep{akbikGeneratingHighQuality2015, aminianTransferringSemanticRoles2017, ebingDevilWordAlignment2025a}. More recently, \citet{chenFrustratinglyEasyLabel2024} investigate joint translation and label projection in one step, inserting square brackets around spans before translation. They report improved downstream performance but degraded translation quality. Subsequent work in the field builds on this finding, separating the translation and label projection steps and applying other techniques such as LLM-based contextual translation or constrained decoding on the unmodified translation \citep{parekhContextualLabelProjection2024, garcia-ferreroTProjectionHighQuality2023, leConstrainedDecodingCrosslingual2024}. While effective, these pipelines introduce considerable computational and engineering overhead.

In this work, we revisit the core assumption motivating these methods, that translation quality is inherently compromised when markers are inserted into the text. We show that with the appropriate training, data, and choice of marker, translation quality can be \textit{improved} while simultaneously transferring labeled spans. 

To this end, we make both a practical and theoretical case for \textit{label-aware translation} with XML tags (\S \ref{sec:labelaware}), and introduce \textbf{LabelPigeon}, a simple approach for joint label projection and translation based on fine-tuning with XML-tagged corpora (\S \ref{sec:labelpigeon}). LabelPigeon conducts both tasks in one pass, handling frequent and nested spans with grace, as we showcase in Figure \ref{fig:labelpigeon}. 

We assess LabelPigeon through three distinct evaluations. We introduce a novel scheme for direct label projection evaluation, verifying LabelPigeon's effectiveness in 11 languages (\S \ref{sec:label}). We further quantify the impact on translation quality across 203 languages as well as varying annotation complexity, finding consistent improvement which we attribute to the additional fine-tuning (\S \ref{sec:translation}). Finally, we conduct downstream experiments on 3 NLP tasks across 27 languages, showcasing that LabelPigeon consistently outperforms prior work, with up to $+40.2$ F1 score improvement (\S \ref{sec:downstream}). Overall, our results indicate that XML tags facilitate effective label projection without compromising translation quality and with no additional computation required at inference, offering a simple alternative to multi-stage pipelines.



\section{Related Work}
\label{sec:related_work}

Several works have explored markup translation in the context of structured-document translation, specifically web pages \citep{bammanTransferringStructuralMarkup2010,joanisTransferringMarkupTags2013, mullerTreatmentMarkupStatistical2017, hannemanHowShouldMarkup2020, hashimotoHighQualityMultilingualDataset2019}. Most rely on a \textit{detag-and-project} approach, where tags are removed, the text translated, and the tags are reinserted \citep{hannemanHowShouldMarkup2020}. 
More recent work investigates the zero-shot capabilities of massively multilingual translation models or large language models (LLMs) on transferring tags, and finds they perform adequately even without any specific fine-tuning \citep{dabreNICTsSubmissionWAT2022, dabreStudyEffectivenessLarge2023, buschbeckMultilingualMultiwayEvaluation2022}. While some works directly train on raw markup data, they exclusively evaluate in the context of structured document translation, largely with translation quality metrics \citep{hannemanHowShouldMarkup2020, hashimotoHighQualityMultilingualDataset2019}. 

Label projection, while sharing structural similarities to markup translation, is largely concerned with transferring annotated span labels for various downstream tasks \citep{chenFrustratinglyEasyLabel2024, ebingTranslateNotTranslate2024}. Alignment-based projection has been widely adopted and is used in projecting data for named entity recognition \citep{niWeaklySupervisedCrossLingual2017}, question answering \citep{huXTREMEMassivelyMultilingual2020, lewisMLQAEvaluatingCrosslingual2020}, event argument extraction \citep{louTranslationBasedImplicitAnnotation2022}, coreference resolution \citep{bitewLazyLowResourceCoreference2021}, and semantic structure \citep{moradshahiLocalizingOpenOntologyQA2020, dazaXSRLParallelCrossLingual2020a}. 
While a subset of prior work utilizes marker-based translation in corpora-building, \citet{chenFrustratinglyEasyLabel2024} are the first to analyze it in depth. They evaluate several marker types in their preliminary zero-shot study, and introduce EasyProject, utilizing synthetically generated data to train a translation model capable of squarebracket-based marker projection. 
Their results indicate a consistent degradation in translation quality, informing later works that opt to separate translation and label projection. 
T-Projection \citep{garcia-ferreroTProjectionHighQuality2023} uses a separate language model to project labels by generating candidate spans, while CLaP \citep{parekhContextualLabelProjection2024} employs a similar approach with an instruction-tuned LLM as a contextual translator. Explicitly motivated by preserving translation quality, \textsc{Codec} \citep{leConstrainedDecodingCrosslingual2024} uses constrained decoding to inject square markers after translation. 
\citet{ebingDevilWordAlignment2025a} further find that word alignment can perform comparably to marker-based label projection with specific low-level design decisions, reinforcing the paradigm of separate label projection. 

Taken together, prior work leaves several aspects unexplored. With the exception of \citet{chenFrustratinglyEasyLabel2024}, no other paper investigates joint translation and label projection. Evaluations rely on indirect metrics such as projection rates or downstream task performance, with no work directly evaluating label projection. In addition, little attention is paid to more challenging cases with frequent, nested, or overlapping spans. Finally, most approaches forgo training altogether or rely on synthetically generated data, leaving existing high-quality data from the field of markup translation underutilized.

\section{Label-Aware Translation}
\label{sec:labelaware}

\begin{figure}[t]
  \includegraphics[width=\columnwidth]{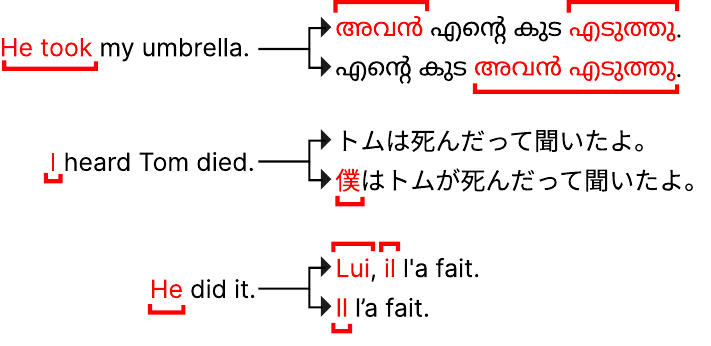}
  \caption{Examples of labeled English sentences with two equally valid translations, where the labeled span is preserved in one and split, omitted, or ambiguous in the other.}
  \label{fig:label_split_examples}
\end{figure}


Prior work assumes that span markers inherently harm translation quality, and therefore designs techniques to project labels on unmarked translations \citep{leConstrainedDecodingCrosslingual2024, garcia-ferreroTProjectionHighQuality2023, parekhContextualLabelProjection2024}. However, we posit that with appropriate training, a label-aware translation is advantageous in several respects. 

Figure \ref{fig:label_split_examples} provides minimal illustrative examples for this purpose. The first example showcases an English sentence that has a labeled span, and two equally valid translations in Malayalam, but one translation preserves the span while the other splits it across the sentence. While splitting the label is not necessarily detrimental, marker-based label projection methods do not have the capability to do so \citep{chenFrustratinglyEasyLabel2024, parekhContextualLabelProjection2024, leConstrainedDecodingCrosslingual2024, garcia-ferreroTProjectionHighQuality2023}, and keeping labeled spans continuous is considered best practice for alignment-based methods as well \citep{ebingDevilWordAlignment2025a}. Similarly, the second example showcases two translations into Japanese, one of which---in an instance of pronoun dropping---omits the label while the other does not. As a highly contextual language, the first translation is generally considered more natural, but the second is also valid, and in our case, preferable. Finally, the third example showcases two translations into French that, depending on the context, can be equally valid. In the first example, the labeled span can arguably be ambiguously assigned to two potential spans: the subject pronoun (``il'') or the stress pronoun (``Lui''). The other, more direct translation is again preferable. 

These examples showcase several potential issues that may arise when translation is done independently of label projection. We hypothesize that joint translation and label transfer would incentivize the model to prioritize the coherence and continuity of the labels.
On the other hand, as illustrated in Figure \ref{fig:label_split_examples}, label-aware translation can also lead to less fluent translations. We argue that less idiomatic translations will typically not lead to substantial annotation quality loss in a model trained on the output data.

\subsection{XML as the Marker of Choice}
\label{subsec:labelaware:xml}

EasyProject, the only prior method utilizing joint label projection and translation, opts for square brackets to mark the label spans, with future work following suit \citep{chenFrustratinglyEasyLabel2024, leConstrainedDecodingCrosslingual2024}.
The authors justify this choice by conducting a preliminary zero-shot study testing out several different markers, with square brackets performing the best. However, this does not translate directly to superior performance after fine-tuning, and the use of square brackets as the marker has several downsides. Most notably, square brackets do not carry direct correspondence between the original spans and the ones in translation. They compensate for this issue with a fuzzy string matching method that translates the annotated spans individually, and matches them to the spans inside the full translation to map the correspondence. This approach is susceptible to errors (in particular when nested or overlapping spans are involved) and balloons inference time as all spans must be translated individually, on top of the text as a whole.

XML tags, on the other hand, provide a direct correspondence between the source spans and translation spans. They can also handle nesting and overlapping spans gracefully, and can even hold semantic information (e.g. \texttt{<PER>} denoting a person) if required. Notably, XML markup has a long history in structured-document translation, with high-quality parallel corpora containing XML-tagged text publicly available \citep{bammanTransferringStructuralMarkup2010, hannemanHowShouldMarkup2020, hashimotoHighQualityMultilingualDataset2019}. In particular, \citet{hashimotoHighQualityMultilingualDataset2019} provide the Salesforce Localization XML MT dataset, a large-scale collection of parallel sentences with naturally occurring XML tags, which can be adapted for training label-aware projection. This resource provides high-quality parallel data that enables models to learn translation while maintaining structured tags, eliminating the need for generating synthetic training data as in prior work \citep{chenFrustratinglyEasyLabel2024}.


\section{LabelPigeon}
\label{sec:labelpigeon}

\begin{figure}[t]
  \includegraphics[width=\columnwidth]{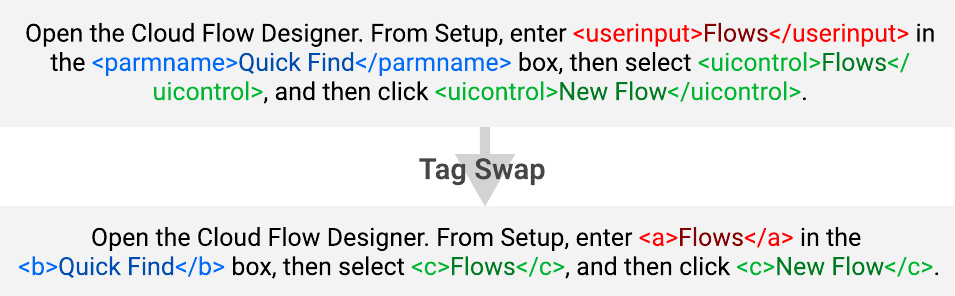}
  \caption{An example showcasing the tag swap that we conduct on training data in order to make it generally applicable.}
  \label{fig:tag-swap}
\end{figure}

Our overarching goal is to re-evaluate the assumption that joint translation and label projection inherently degrades quality. As such, we focus on the effects of direct fine-tuning with high-quality data, and to that end, we opt for the Salesforce Localization XML MT dataset mentioned in \S \ref{subsec:labelaware:xml} \citep{hashimotoHighQualityMultilingualDataset2019}. A gold-standard XML-tagged corpus, it consists of parallel pairs between English and seven other languages with approximately 100,000 aligned samples in each language pair, providing ample data for full-scale fine-tuning. 

Prior research indicates that fine-tuning on too large of a dataset or on low-resource languages could lead to catastrophic forgetting and a general reduction in translation quality \citep{liuConditionsCatastrophicForgetting2025, chenFrustratinglyEasyLabel2024}. We further conduct ablation experiments (detailed in Appendix \ref{sec:appendix:ablations}) and find that fine-tuning on all seven language pairs is counterproductive. In accordance, we opt to train the final model with data between English and three high-resource languages: German, Russian, and Chinese.

The tags are largely composed of UI and styling elements. In order to adapt the dataset for general label projection, we opt to swap these for simple alphabetical non-descript tags of the form \texttt{<a>}, \texttt{<b>}, etc. All tags of a certain type are converted into a corresponding alphabetical tag based on the order of appearance. Figure \ref{fig:tag-swap} showcases an example of this in action. We also drop all examples that contain no tags, resulting in a sizeable reduction of the dataset to 25k samples in each language pair. 
Across all three datasets, and accounting for translation in both directions with English, this amounts to approximately 150k training samples, of which 5\% is utilized as a development set. 

Due to its effectiveness, coverage, and widespread use, we opt for the NLLB-200 3.3B as the base translation model to fine-tune \citep{teamNoLanguageLeft2022}.  We conduct fine-tuning on our modified dataset for a full epoch, totaling 9,091 steps with an effective batch size of 16, taking 5h:30m on a single NVIDIA A100 GPU. Additional training and data specifics are given in Appendix \ref{sec:appendix:training}.

With this model, label projection can be conducted in a straightforward procedure: insert alphabetical XML tags on the annotated spans, translate with our model, and extract the tags using an off-the-shelf XML parser. We term our method \textbf{LabelPigeon}, and note that it has a negligible computational overhead at inference, requiring only a single forward pass of the model.

\section{Directly Evaluating Label Projection}
\label{sec:label}

\begin{table*}[t]
  \centering
  \begin{tabular}{l r r r r r r r r}
  \toprule
  \multirow[c]{2}{*}{\textbf{Language}} & \multicolumn{4}{c}{\textbf{COMET Score}} & \multicolumn{4}{c}{\textbf{Label Match F1 (\%)}} \\
  \cmidrule(lr){2-5} \cmidrule(lr){6-9}
   & \textbf{Awes.} & \textbf{Gemma} & \textbf{EProj.} & \textbf{LP} & \textbf{Awes.} & \textbf{Gemma} & \textbf{EProj.} & \textbf{LP} \\
  \midrule
  \multicolumn{9}{l}{\textbf{XQUAD}} \\
  \cmidrule(lr){1-9}
  Arabic & \textbf{\textcolor{black!85}{83.4}} & \cellcolor{BrickRed!25}54.3 & \cellcolor{BrickRed!9}81.3 & \cellcolor{BrickRed!5}82.2 & 49.8 & \cellcolor{ForestGreen!20}69.8 & \cellcolor{ForestGreen!31}\textbf{\textcolor{black!85}{80.9}} & \cellcolor{ForestGreen!26}75.3 \\
  Chinese & \textbf{\textcolor{black!85}{80.8}} & \cellcolor{BrickRed!25}64.9 & \cellcolor{BrickRed!6}79.4 & \cellcolor{BrickRed!12}77.9 & 46.4 & \cellcolor{ForestGreen!24}70.6 & \cellcolor{ForestGreen!24}70.9 & \cellcolor{ForestGreen!26}\textbf{\textcolor{black!85}{72.9}} \\
  German & \textbf{\textcolor{black!85}{83.0}} & \cellcolor{BrickRed!25}73.7 & \cellcolor{BrickRed!8}81.1 & \cellcolor{BrickRed!1}82.7 & 60.0 & \cellcolor{ForestGreen!26}86.4 & \cellcolor{ForestGreen!25}84.6 & \cellcolor{ForestGreen!27}\textbf{\textcolor{black!85}{86.8}} \\
  Greek & 84.2 & \cellcolor{BrickRed!25}71.4 & \cellcolor{ForestGreen!0}84.3 & \cellcolor{ForestGreen!12}\textbf{\textcolor{black!85}{87.3}} & 44.8 & \cellcolor{ForestGreen!27}71.4 & \cellcolor{ForestGreen!21}65.8 & \cellcolor{ForestGreen!31}\textbf{\textcolor{black!85}{75.8}} \\
  Hindi & \textbf{\textcolor{black!85}{78.4}} & \cellcolor{BrickRed!25}50.1 & \cellcolor{BrickRed!8}76.4 & \cellcolor{BrickRed!5}77.1 & 54.8 & \cellcolor{ForestGreen!16}71.3 & \cellcolor{ForestGreen!28}\textbf{\textcolor{black!85}{82.6}} & \cellcolor{ForestGreen!22}76.9 \\
  Romanian & 84.3 & \cellcolor{BrickRed!2}83.8 & \cellcolor{BrickRed!2}83.9 & \cellcolor{ForestGreen!7}\textbf{\textcolor{black!85}{86.0}} & 58.0 & \cellcolor{ForestGreen!31}\textbf{\textcolor{black!85}{89.2}} & \cellcolor{ForestGreen!24}81.6 & \cellcolor{ForestGreen!30}87.8 \\
  Russian & 83.7 & \cellcolor{BrickRed!19}79.0 & \cellcolor{BrickRed!3}82.8 & \cellcolor{ForestGreen!5}\textbf{\textcolor{black!85}{85.0}} & 52.5 & \cellcolor{ForestGreen!29}\textbf{\textcolor{black!85}{81.8}} & \cellcolor{ForestGreen!24}76.6 & \cellcolor{ForestGreen!26}78.9 \\
  Spanish & 83.1 & \cellcolor{BrickRed!25}68.8 & \cellcolor{BrickRed!6}81.7 & \cellcolor{ForestGreen!4}\textbf{\textcolor{black!85}{84.0}} & 59.2 & \cellcolor{ForestGreen!29}87.8 & \cellcolor{ForestGreen!23}82.4 & \cellcolor{ForestGreen!31}\textbf{\textcolor{black!85}{90.2}} \\
  Thai & \textbf{\textcolor{black!85}{76.8}} & \cellcolor{BrickRed!25}67.8 & \cellcolor{BrickRed!11}74.1 & \cellcolor{BrickRed!1}76.6 & 23.8 & \cellcolor{ForestGreen!41}64.9 & \cellcolor{ForestGreen!42}\textbf{\textcolor{black!85}{66.0}} & \cellcolor{ForestGreen!39}63.1 \\
  Turkish & 84.0 & \cellcolor{BrickRed!21}78.9 & \cellcolor{BrickRed!4}83.0 & \cellcolor{ForestGreen!4}\textbf{\textcolor{black!85}{85.0}} & 58.4 & \cellcolor{ForestGreen!26}\textbf{\textcolor{black!85}{84.9}} & \cellcolor{ForestGreen!26}84.0 & \cellcolor{ForestGreen!25}83.3 \\
  Vietnamese & 83.1 & \cellcolor{BrickRed!25}72.6 & \cellcolor{BrickRed!9}80.8 & \cellcolor{ForestGreen!1}\textbf{\textcolor{black!85}{83.3}} & 48.8 & \cellcolor{ForestGreen!32}\textbf{\textcolor{black!85}{80.9}} & \cellcolor{ForestGreen!30}78.8 & \cellcolor{ForestGreen!31}79.7 \\
  \textbf{\textcolor{black!85}{Average}} & 82.3 & \cellcolor{BrickRed!25}69.6 & \cellcolor{BrickRed!6}80.8 & \cellcolor{ForestGreen!1}\textbf{\textcolor{black!85}{82.4}} & 50.6 & \cellcolor{ForestGreen!27}78.1 & \cellcolor{ForestGreen!27}77.7 & \cellcolor{ForestGreen!29}\textbf{\textcolor{black!85}{79.2}} \\
  \midrule
  \multicolumn{9}{l}{\textbf{MLQA}} \\
  \cmidrule(lr){1-9}
  Arabic & \textbf{\textcolor{black!85}{84.8}} & \cellcolor{BrickRed!25}47.2 & \cellcolor{BrickRed!5}83.7 & \cellcolor{BrickRed!0}84.8 & 51.9 & \cellcolor{ForestGreen!13}65.0 & \cellcolor{ForestGreen!29}\textbf{\textcolor{black!85}{80.7}} & \cellcolor{ForestGreen!26}78.0 \\
  Chinese & \textbf{\textcolor{black!85}{80.4}} & \cellcolor{BrickRed!25}53.7 & \cellcolor{BrickRed!5}79.1 & \cellcolor{BrickRed!3}79.6 & 40.6 & \cellcolor{ForestGreen!12}52.9 & \cellcolor{ForestGreen!23}63.9 & \cellcolor{ForestGreen!27}\textbf{\textcolor{black!85}{67.8}} \\
  German & 82.4 & \cellcolor{BrickRed!25}59.9 & \cellcolor{BrickRed!4}81.4 & \cellcolor{ForestGreen!6}\textbf{\textcolor{black!85}{84.0}} & 60.3 & \cellcolor{ForestGreen!18}77.9 & \cellcolor{ForestGreen!17}77.2 & \cellcolor{ForestGreen!23}\textbf{\textcolor{black!85}{83.4}} \\
  Hindi & 76.7 & \cellcolor{BrickRed!25}45.3 & \cellcolor{BrickRed!4}75.7 & \cellcolor{ForestGreen!1}\textbf{\textcolor{black!85}{76.9}} & 56.1 & \cellcolor{ForestGreen!7}63.4 & \cellcolor{ForestGreen!24}\textbf{\textcolor{black!85}{80.1}} & \cellcolor{ForestGreen!23}79.3 \\
  Spanish & 82.5 & \cellcolor{BrickRed!25}59.1 & \cellcolor{BrickRed!1}82.3 & \cellcolor{ForestGreen!6}\textbf{\textcolor{black!85}{84.0}} & 58.6 & \cellcolor{ForestGreen!18}76.3 & \cellcolor{ForestGreen!20}78.9 & \cellcolor{ForestGreen!30}\textbf{\textcolor{black!85}{88.8}} \\
  Vietnamese & 83.0 & \cellcolor{BrickRed!25}62.2 & \cellcolor{BrickRed!1}82.7 & \cellcolor{ForestGreen!6}\textbf{\textcolor{black!85}{84.7}} & 49.1 & \cellcolor{ForestGreen!24}73.0 & \cellcolor{ForestGreen!29}78.5 & \cellcolor{ForestGreen!33}\textbf{\textcolor{black!85}{82.3}} \\
  \textbf{\textcolor{black!85}{Average}} & 81.6 & \cellcolor{BrickRed!25}54.6 & \cellcolor{BrickRed!3}80.8 & \cellcolor{ForestGreen!3}\textbf{\textcolor{black!85}{82.3}} & 52.8 & \cellcolor{ForestGreen!15}68.1 & \cellcolor{ForestGreen!24}76.5 & \cellcolor{ForestGreen!27}\textbf{\textcolor{black!85}{79.9}} \\
  
  \bottomrule
\end{tabular}
\caption{Direct label projection results on XQuAD and MLQA. COMET scores and the label match F1 scores are both provided. Sentences are translated from English to the corresponding language. We compare four label projection methods: a) Awesome-align (Awes.), b) Gemma 3 27B (Gemma), c) EasyProject (EProj.), and d) LabelPigeon (LP). Awesome-align is used as the baseline, and differences are highlighted via color.}
\label{tab:direct_evals}
\end{table*}

\begin{table*}[t]
  \centering
  \begin{tabular}{l r r r r r r r r r r}
    \toprule
    \multirow[c]{2}{*}{\textbf{Metrics}} & \multicolumn{4}{c}{\textbf{No Markers}} & \multicolumn{2}{c}{\textbf{Single}} & \multicolumn{2}{c}{\textbf{Simple}} & \multicolumn{2}{c}{\textbf{Complex}} \\
    \cmidrule(lr){2-5}\cmidrule(lr){6-7}\cmidrule(lr){8-9}\cmidrule(lr){10-11}
     & \textbf{Baseline} & \textbf{EProj.} & \textbf{LP} & \textbf{NF} & \textbf{EProj.} & \textbf{LP} & \textbf{EProj.} & \textbf{LP} & \textbf{EProj.} & \textbf{LP} \\
    \midrule
    BLEU & 17.4 & \cellcolor{ForestGreen!5}17.7 & \cellcolor{ForestGreen!3}17.6 & \cellcolor{ForestGreen!8}17.9 & \cellcolor{BrickRed!6}16.8 & \cellcolor{ForestGreen!3}17.6 & \cellcolor{BrickRed!23}15.3 & \cellcolor{BrickRed!14}16.1 & \cellcolor{BrickRed!28}14.9 & \cellcolor{BrickRed!21}15.5 \\
    chrF++ & 42.9 & \cellcolor{ForestGreen!14}43.5 & \cellcolor{ForestGreen!11}43.4 & \cellcolor{ForestGreen!20}43.8 & \cellcolor{BrickRed!0}42.8 & \cellcolor{ForestGreen!18}43.7 & \cellcolor{BrickRed!20}41.4 & \cellcolor{BrickRed!7}42.3 & \cellcolor{BrickRed!28}40.8 & \cellcolor{BrickRed!16}41.7 \\
    Proj. Rate & — & — & — & — & 85.9 & 92.5 & 68.2 & 81.0 & 47.7 & 69.3 \\
    \bottomrule
  \end{tabular}

  \caption{Metrics across different models and with different marker-insertion strategies on the \textsc{Flores-200} dataset. We compare the baseline NLLB-200 3.3B model (Baseline), EasyProject (EProj.), LabelPigeon (LP), and the Non-marker Fine-tuned model (NF). BLEU and chrF++ are lexical measures for translation quality, while the projection rate measures how many of the original labels are included in the translation. Differences with respect to the baseline are highlighted via color.}
  \label{tab:flores}
\end{table*}

Prior work generally evaluates label projection methods by translating span-annotated datasets and training models on those datasets, essentially using the downstream results as proxy for the efficacy of the label projection \citep{chenFrustratinglyEasyLabel2024, garcia-ferreroTProjectionHighQuality2023, parekhContextualLabelProjection2024, leConstrainedDecodingCrosslingual2024}. We instead opt to define our own benchmark and metrics to directly evaluate label projection, utilizing parallel span-annotated datasets.

\subsection{Experimental Setup}
\label{subsec:label:setup}

\paragraph{Datasets.} For directly evaluating label projection we utilize XQuAD \citep{artetxeCrosslingualTransferabilityMonolingual2020} and MLQA \citep{lewisMLQAEvaluatingCrosslingual2020}, two gold-standard multilingual extractive question-answering (QA) datasets. XQuAD consists of 240 paragraphs and 1190 QA pairs in 12 languages, with the other 11 languages translated from English, while MLQA consists of over 5,000 QA pairs in 7 languages that were mined from Wikipedia. Both datasets provide span-annotated QA data parallel across multiple languages, allowing direct measurement of how well projected spans align in the target language. Because MLQA is not consistently parallel at the paragraph level, we apply a simple filter to only retain the QA pairs with parallel contexts, detailed in Appendix \ref{sec:appendix:mlqa}. Additional statistics, particularly with regards to label frequency, are provided in Appendix \ref{sec:appendix:label}.


\paragraph{Metrics.} Across both datasets, we define a simple evaluation scheme to concretely evaluate the accuracy of label projection. Each projected label span is taken individually and is considered a match if it has string similarity above a set ratio to the corresponding reference label span, where similarity is computed with Ratcliff/Obershelp pattern matching \citep{black2004ratcliff}. Our main metric is the global F1 score of these label matches, which we refer to as \textbf{Label Match F1}. For all our experiments, we set the aforementioned string similarity ratio to 50\%. We also calculate the COMET score (specifically COMET-22, \citealp{reiCOMET22UnbabelIST20222022}) to evaluate the impact on translation quality. 
We note that all markers are removed before evaluating translation quality. 

\paragraph{Baselines.} We compare LabelPigeon with the following baselines: (1) Awesome-align \citep{douWordAlignmentFinetuning2021a}, an alignment-based label projection method; (2) Gemma 3 27B IT \citep{teamGemma3Technical2025}, a strong lightweight open-source LLM; and (3) EasyProject \citep{chenFrustratinglyEasyLabel2024}, a marker-based label projection method. As Awesome-align conducts label projection separately after translation, we opt for the original NLLB-200 3.3B as the corresponding translation model, providing direct comparison in terms of translation. For EasyProject, we use their fine-tuned NLLB-200 3.3B model for the same reason.  Additional details can be found in Appendix \ref{sec:appendix:label}.

\subsection{Results}
\label{subsec:label:results}

Table \ref{tab:direct_evals} compiles the label projection results across languages and models, showcasing both the COMET scores and Label Match F1.   
We first note that our method outperforms all other baselines in label projection. 
Awesome-align performs particularly poorly, with an average label match F1 of $50.6/52.8$ on XQuAD/MLQA. EasyProject and Gemma 3 perform reasonably well with average F1 scores of $77.7/76.5$ and $78.1/68.1$ respectively, but are still outperformed by LabelPigeon's $79.2/79.9$. We also note that the training dataset contains a maximum of 6 unique tags per example (i.e. up to \texttt{<f>}). In contrast, XQuAD samples contain more than 9 tags on average, with a maximum of 24 tags (up to \texttt{<x>}). Given the performant results of LabelPigeon, we conclude that it is able to generalize up to much higher unique tag counts than seen during training. 

The translation quality results warrant closer scrutiny. While EasyProject degrades translation quality across the board as we expect, our method \textit{improves} translation quality over the base NLLB model for a majority of languages. We note that this is equally true for languages that we did not fine-tune on, as well as the three that we did (German, Chinese, and Russian). We further explore the cause for this improvement in \S \ref{sec:translation}. Gemma 3, the only method not utilizing a base or fine-tuned version of NLLB-200 3.3B, provides markedly poorer translations than all other baselines over almost all languages, showcasing the need for translation-specific models for the task.

Additionally, we perform a small-scale error analysis on the XQuAD data for our method. We manually annotate a random sample of 30 examples translated from English to German via LabelPigeon, and compare it with the ground truth for translation errors in the labels.
Out of 137 total labels, 118 were considered correct, resulting in a span translation accuracy of 86\%. With respect to the automatic evaluation, we observed only three false positives and five false negatives. These findings further reinforce the effectiveness of our automated evaluation. 

\section{Impact on Translation Quality}
\label{sec:translation}

While we jointly evaluate translation quality and label projection in \S\ref{sec:label}, we only cover 11 languages. Additionally, the phenomenon of improved translation quality with our method warrants further investigation. In order to provide a more comprehensive overview of how markers and training impact translation quality, we opt for a broad-scale evaluation with synthetically inserted markers on the \textsc{Flores-200} dataset \citep{teamNoLanguageLeft2022}. 

\subsection{Experimental Setup}
\label{subsec:translation:setup}

\paragraph{Dataset.} \textsc{Flores-200} is an extension of the well-known \textsc{Flores-101} \citep{goyalFlores101EvaluationBenchmark2022}, expanding it to cover 204 languages, and it was extensively used by \citet{teamNoLanguageLeft2022} to evaluate the NLLB-200 model. We use the publicly available \texttt{devtest} split containing 1012 sentences, and evaluate translation quality from English to all 203 other languages. 

\paragraph{Synthetic Markers.} As the dataset itself does not contain any sort of labeled spans, we simulate labels by randomly inserting markers on the English source sentences, in various configurations. Specifically, we utilize three marker insertion configurations representing different labelling scenarios: the \textbf{Single} configuration always inserts exactly one marker, the \textbf{Simple} configuration inserts non-overlapping and non-nested markers, and the \textbf{Complex} configuration inserts potentially overlapping and nested markers. The specific algorithm is elaborated in Appendix \ref{sec:appendix:synthetic_marker}. We also test on the original unmarked data, referred to as the \textbf{No Markers} configuration.

\paragraph{Metrics.} Following \citet{teamNoLanguageLeft2022}, we opt for lexical measures of translation quality, specifically BLEU \citep{papineniBleuMethodAutomatic2002} and chrF++ \citep{popovicChrFWordsHelping2017}. We additionally measure label projection via the projection rate, defined by \citet{chenFrustratinglyEasyLabel2024} as the percentage of data in which the numbers and type of special markers in the translations match with the source sentences. We note that this metric only takes into consideration the existence of the markers and not the accuracy of the labels themselves, and thus is significantly less reliable than our direct label projection schema in \S \ref{sec:label}.

\paragraph{Baselines.} As we are largely concerned with the impact of training on translation quality, we compare NLLB 3.3B with models derived from it, mainly LabelPigeon (LP) and EasyProject (EProj). Additionally, to disambiguate the effect of additional training with the effects of marker insertion itself, we train a model on the modified SalesForce Localization XML MT dataset as in \S \ref{sec:labelpigeon}, but with the XML tags removed. All other hyperparameters are kept the same, and we refer to this model as the \textbf{Non-marker Fine-tuned} (NF) model.

\subsection{Results}
\label{subsec:translation:results}

The results are compiled in Table \ref{tab:flores}. We start by noting the improvement in translation quality of all three fine-tuned models over the baseline model when no markers are inserted. As markers are introduced, translation quality degrades for EasyProject, both compared to itself and the baseline in the No Marker configuration. However, the BLEU score remains the same, and chrF++ increases when a single marker per sentence is inserted for LabelPigeon. With multiple and nested markers, we see a clear decline in translation quality for both EasyProject and LabelPigeon. Regardless, LabelPigeon consistently outperforms EasyProject across all marker insertion configurations in both translation quality metrics. In addition, LabelPigeon attains a higher projection rate across the board, while EasyProject struggles particularly in the Complex marker insertion scheme. Taken in conjunction with the results from \S \ref{sec:label}, we can confidently state that LabelPigeon improves translation quality when used on span-marked data. We provide the full results in Appendix \ref{sec:appendix:full_flores_results} and additional experiments on the effects of length and frequency in Appendix \ref{sec:appendix:marker_freq}.

\paragraph{Why Does Translation Quality Improve?} 
The performance of the NF model, which has been trained on unmarked data and performs the best overall, shows that the quality improvement is a direct result of additional training.
This is consistent with prior research that show fine-tuning translation models on small datasets (approx. 100K sentences) can induce positive cross-lingual transfer, improving performance for even unseen languages  \citep{liuConditionsCatastrophicForgetting2025}. Regardless, LabelPigeon's performance with single markers is comparable to the NF model's performance under no markers, providing evidence for our hypothesis in \S \ref{sec:labelaware}: that the less idiomatic translations resulting from label-aware translation does not lead to a substantial quality loss.

\section{Downstream Experiments}
\label{sec:downstream}

\begin{table}[btp]
  \centering
  \small
  \begin{tabular}{l l r r}
    \toprule
    \textbf{Language} & \textbf{Dataset} & \textbf{EProj.} & \textbf{Ours} \\
    \midrule
    \multicolumn{4}{l}{\textbf{UNER (Named Entity Recognition)}} \\
    \midrule
    Cebuano & \texttt{ceb\_gja} & 47.6 & \textbf{\cellcolor{ForestGreen!39}78.3} \\
    \addlinespace
    \multirow[c]{3}{*}{Chinese} & \texttt{zh\_gsd} & \textbf{53.9} & \cellcolor{BrickRed!6}46.4 \\
     & \texttt{zh\_gsdsimp} & \textbf{52.9} & \cellcolor{BrickRed!5}47.4 \\
     & \texttt{zh\_pud} & \textbf{62.2} & \cellcolor{BrickRed!6}54.5 \\
    \addlinespace
    Croatian & \texttt{hr\_set} & 77.4 & \textbf{\cellcolor{ForestGreen!10}85.6} \\
    Danish & \texttt{da\_ddt} & 75.5 & \textbf{\cellcolor{ForestGreen!5}79.3} \\
    German & \texttt{de\_pud} & 76.9 & \textbf{\cellcolor{ForestGreen!4}80.2} \\
    \addlinespace
    \multirow[c]{2}{*}{Portuguese} & \texttt{pt\_bosque} & 62.2 & \textbf{\cellcolor{ForestGreen!27}83.0} \\
     & \texttt{pt\_pud} & 65.1 & \textbf{\cellcolor{ForestGreen!27}86.1} \\
    \addlinespace
    Russian & \texttt{ru\_pud} & 56.7 & \textbf{\cellcolor{ForestGreen!17}70.4} \\
    Serbian & \texttt{sr\_set} & 74.8 & \textbf{\cellcolor{ForestGreen!16}87.4} \\
    Slovak & \texttt{sk\_snk} & 64.3 & \textbf{\cellcolor{ForestGreen!18}78.6} \\
    \addlinespace
    \multirow[c]{2}{*}{Swedish} & \texttt{sv\_pud} & 70.8 & \textbf{\cellcolor{ForestGreen!22}87.7} \\
     & \texttt{sv\_talbanken} & 67.3 & \textbf{\cellcolor{ForestGreen!27}88.5} \\
    \addlinespace
    \multirow[c]{2}{*}{Tagalog} & \texttt{tl\_trg} & 54.1 & \textbf{\cellcolor{ForestGreen!48}91.5} \\
     & \texttt{tl\_ugnayan} & 38.5 & \textbf{\cellcolor{ForestGreen!55}81.5} \\
     \addlinespace
    \textbf{Average} & -- & 62.5 & \textbf{\cellcolor{ForestGreen!18}76.7} \\
    \midrule
    \multicolumn{4}{l}{\textbf{CorefUD (Coreference Resolution)}} \\
    \midrule
    Ancient Greek & PROIEL & \cellcolor{Gray!50}0.0 & \cellcolor{Gray!50}0.0 \\
    Ancient Hebrew & PTNK & \cellcolor{Gray!50}0.0 & \cellcolor{Gray!50}0.0 \\
    Catalan & AnCora & 1.5 & \textbf{\cellcolor{ForestGreen!18}12.1} \\
    \addlinespace
    \multirow[c]{2}{*}{Czech} & PCEDT & 1.8 & \textbf{\cellcolor{ForestGreen!32}20.5} \\
     & PDT & \cellcolor{Gray!50}0.5 & \textbf{\cellcolor{ForestGreen!34}20.4} \\
    \addlinespace
    \multirow[c]{2}{*}{French} & ANCOR & \cellcolor{Gray!50}0.2 & \textbf{\cellcolor{ForestGreen!5}3.4} \\
     & Democrat & \cellcolor{Gray!50}0.1 & \textbf{\cellcolor{ForestGreen!2}1.2} \\
    \addlinespace
    \multirow[c]{2}{*}{German} & ParCorFull & \textbf{18.7} & \cellcolor{BrickRed!6}12.7 \\
     & PotsdamCC & \textbf{19.5} & \cellcolor{BrickRed!4}16.2 \\
    \addlinespace
    Hindi & HDTB & \cellcolor{Gray!50}0.0 & \textbf{\cellcolor{ForestGreen!46}27.2} \\
    \addlinespace
    \multirow[c]{2}{*}{Hungarian} & KorKor & \cellcolor{Gray!50}0.0 & \textbf{\cellcolor{ForestGreen!6}3.8} \\
     & SzegedKoref & \cellcolor{Gray!50}0.0 & \textbf{\cellcolor{ForestGreen!4}2.3} \\
    \addlinespace
    Korean & ECMT & \cellcolor{Gray!50}0.0 & \textbf{\cellcolor{ForestGreen!11}6.3} \\
    Lithuanian & LCC & \cellcolor{Gray!50}0.0 & \textbf{\cellcolor{ForestGreen!43}25.5} \\
    \addlinespace
    \multirow[c]{2}{*}{Norwegian} & BokmaalNARC & \cellcolor{Gray!50}0.1 & \textbf{\cellcolor{ForestGreen!53}31.6} \\
     & NynorskNARC & \cellcolor{Gray!50}0.2 & \textbf{\cellcolor{ForestGreen!55}32.8} \\
    \addlinespace
    Old Slavonic & PROIEL & \cellcolor{Gray!50}0.4 & \textbf{\cellcolor{ForestGreen!2}1.7} \\
    Polish & PCC & 4.1 & \textbf{\cellcolor{ForestGreen!13}12.0} \\
    Russian & RuCor & 10.2 & \textbf{\cellcolor{ForestGreen!38}32.7} \\
    Spanish & AnCora & \cellcolor{Gray!50}0.4 & \textbf{\cellcolor{ForestGreen!17}10.6} \\
    Turkish & ITCC & \cellcolor{Gray!50}0.1 & \textbf{\cellcolor{ForestGreen!21}12.4} \\
    \addlinespace
    \textbf{Average} & -- & 2.7 & \textbf{\cellcolor{ForestGreen!18}13.6} \\
    \midrule
    \multicolumn{4}{l}{\textbf{MLQA (Question Answering)}} \\
    \midrule
    Arabic & -- & 62.6 & \textbf{\cellcolor{ForestGreen!2}62.7} \\
    Chinese & -- & \textbf{53.5} & \cellcolor{BrickRed!2}53.4 \\
    German & -- & 65.6 & \textbf{\cellcolor{ForestGreen!27}67.5} \\
    Hindi & -- & \textbf{70.8} & \cellcolor{BrickRed!6}69.7 \\
    Spanish & -- & 71.4 & \textbf{\cellcolor{ForestGreen!11}72.2} \\
    Vietnamese & -- & \textbf{72.9} & \cellcolor{BrickRed!13}71.5 \\
    \midrule
    \textbf{Average} & -- & 66.1 & \textbf{\cellcolor{ForestGreen!3}66.1} \\

    \bottomrule
  \end{tabular}
  \caption{Downstream F1 scores for UNER, CorefUD, and MLQA, comparing EasyProject (EProj.) and LabelPigeon (Ours). Differences are highlighted via color, and instances of exceptionally low scores (F1 $<1$) are noted in gray.}
  \label{tab:downstream}
\end{table}


In line with recent work and prior applications, we evaluate the effectiveness of our label projection method on three downstream tasks: named entity recognition (NER), question answering (QA), and coreference resolution (CR) \citep{bitewLazyLowResourceCoreference2021, chenFrustratinglyEasyLabel2024}.

\subsection{Experimental Setup}
\label{subsec:downstream:setup}

\paragraph{Named Entity Recognition.} 
To evaluate NER, we opt for Universal Named Entity Recognition (UNER), a recently released gold-standard benchmark containing 19 datasets across 13 diverse languages \citep{mayhewUniversalNERGoldStandard2024}.  
We use the training split of the English portion of the dataset (i.e., the EWT dataset) as the source for cross-lingual transfer. In line with their baseline, we train XLM-R\textsubscript{Large} (560M parameters) on the translated data, and use the test splits of all other languages and corresponding datasets for evaluation \citep{conneauUnsupervisedCrosslingualRepresentation2020}.

\paragraph{Question Answering.} As in \S \ref{subsec:label:setup}, we use MLQA \citep{lewisMLQAEvaluatingCrosslingual2020} for question-answering evaluation. Due to the comparatively smaller number of evaluation samples, we omit XQuAD for a downstream comparison. We use SQuAD v1.1 \citep{rajpurkarSQuAD100000Questions2016} as the source dataset, and again opt for XLM-R\textsubscript{Large} as it is the best-performing baseline on MLQA, with F1 scores as the metric.

\paragraph{Coreference Resolution.} For coreference resolution, we use the publicly available version of the widely known CorefUD 1.3 dataset, covering 24 datasets in 17 languages \citep{nedoluzhkoCorefUD10Coreference2022, corefud13}. While \citet{nedoluzhkoCorefUD10Coreference2022} provide no baseline, the CRAC shared task for multilingual coreference resolution utilizes multilingual BERT\textsubscript{base} \citep{devlinBERTPretrainingDeep2019} as their baseline, which we adopt \citep{prazakMultilingualCoreferenceResolution2021, novakFindingsThirdShared2024}. We use the English portion of OntoNotes 5.0 (\citealp{weischedelralphOntoNotesRelease502013}, also part of CorefUD) as the source dataset to translate, and evaluate on all other languages and corresponding datasets, of which there are 21.

\paragraph{Baselines.} Given the subpar performance of Gemma 3 and Awesome-align in our direct label projection results (\S \ref{subsec:label:results}), and the compute costs associated with translating large datasets, we opt to only evaluate EasyProject and LabelPigeon for the downstream experiments. Training hyperparameters and other specifics are provided in Appendix \ref{sec:appendix:downstream}.

\subsection{Results}
\label{subsec:downstream:results}

The results on each of the component tasks and datasets are compiled in Table \ref{tab:downstream}. Through all three tasks, LabelPigeon outperforms EasyProject in the majority of datasets. For NER, we see large and consistent gains across most languages, with an average improvement of $+14.2$ and particularly significant improvements in low-resource languages such as Cebuano ($+30.7$) and Tagalog ($+40.2$). LabelPigeon also generally provides strong downstream performance with F1 scores above $80$ in the majority of datasets. 

In contrast, performance remains low across the board for coreference resolution. We hypothesize that this is due to a combination of two factors: the inherent frequency and nesting of coreference spans, which we have shown to reduce translation quality and label projection accuracy (\S \ref{sec:translation}), and the overall difficulty of the task, as reflected in the low average baseline score of $54.75$ even with high-quality in-domain training data \citep{novakFindingsThirdShared2024}. For certain languages such as Ancient Greek and Ancient Hebrew, the downstream model fails to annotate at all, resulting in scores of $0.0$. However, this phenomenon of complete failure occurs much more frequently for EasyProject than for LabelPigeon. Across the 16 languages tested, EasyProject yields scores $<1.0$ in 11 of them, while LabelPigeon only fails by this criterion for the two aforementioned historical languages. The only languages where EasyProject obtains somewhat functional results are German with $19.1$ and Russian with $10.2$. In contrast, LabelPigeon achieves $>10.0$ for 10 languages, and $>20.0$ for 5. 

Finally, for question answering, we observe only a narrow gap between LabelPigeon at $66.15$ and EasyProject at $66.13$, with both methods performing comparably well. Nevertheless, LabelPigeon still outperforms EasyProject across all three tasks on average, consistent with the results from the direct evaluation in \S \ref{sec:label}.

\section{Conclusion}

In this work, we present the case for joint label projection and translation with XML tags as the marker of choice.
Through comprehensive evaluations covering direct label projection, translation quality, and downstream effectiveness, we show that our method outperforms existing marker-based and alignment-based methods without incurring engineering overhead or additional computation at inference. In the broader context of a field that has largely abandoned this approach in favor of complex multi-stage pipelines, our work shows that a straightforward training regimen and high-quality data can provide effective label projection without harming translation quality. 




\section*{Limitations}


The direct label projection evaluation as detailed in \S \ref{sec:label} utilizes XQuAD, where all samples are translated from English, and a filtered version of MLQA, where the filtering may bias it towards direct translations. We also use \textsc{Flores-200}, another directly translated dataset, in \S \ref{sec:translation} to evaluate translation quality. As such, these evaluations may be affected by the phenomenon of \textit{translationese}, where human-translated text can contain unusual features not present in natural text \citep{grahamTranslationeseMachineTranslation2019, bakerCorpusLinguisticsTranslation1993}. 

While we use three different tasks for downstream evaluation, we only use question answering datasets for the direct label evaluation, largely composed of high-resource languages. However, due to the requirements of such an evaluation (namely needing to be multilingual, parallel, and span-annotated), very few datasets are fit for this purpose. 
Our synthetic tag insertion in \S \ref{sec:translation} may also not accurately reflect real-world usage, as tags are typically motivated by semantics or linguistics. Regardless, our results on it are consistent with the translation quality improvements observed in our direct label evaluation.

Finally, we do not conduct a full evaluation of the newest label projection methods such as \textsc{Codec} \citep{leConstrainedDecodingCrosslingual2024} and CLaP \citep{parekhContextualLabelProjection2024}. In preliminary experiments, we found that \textsc{Codec} was outperformed by LabelPigeon, and a full evaluation was prohibitively expensive, as we describe in Appendix \ref{sec:appendix:codec}. We make the case for and focus on label-aware translation, and given the extensive engineering and additional inference requirements of these methods, we leave their exploration to future work.


\section*{Ethical Considerations}
Label projection has the potential to bring higher-quality labels to low resource languages.
While this is generally a worthwhile pursuit, one might argue that culturally sensitive annotations that cover specific linguistic phenomena are disincentivized by better label projection.
We argue that these risks are outweighed by the benefit of more accessible NLP for lower resource languages. Overall, we do not anticipate major ethical concerns arising from this work.
 

\bibliography{custom}

@inproceedings{bitewLazyLowResourceCoreference2021,
  title = {Lazy {{Low-Resource Coreference Resolution}}: A {{Study}} on {{Leveraging Black-Box Translation Tools}}},
  shorttitle = {Lazy {{Low-Resource Coreference Resolution}}},
  booktitle = {Proceedings of the {{Fourth Workshop}} on {{Computational Models}} of {{Reference}}, {{Anaphora}} and {{Coreference}}},
  author = {Bitew, Semere Kiros and Deleu, Johannes and Develder, Chris and Demeester, Thomas},
  year = {2021},
  pages = {57--62},
  publisher = {Association for Computational Linguistics},
  address = {Punta Cana, Dominican Republic},
  doi = {10.18653/v1/2021.crac-1.6},
  urldate = {2025-09-09},
  langid = {english}
}

@inproceedings{mayhewUniversalNERGoldStandard2024,
  title = {Universal {{NER}}: {{A Gold-Standard Multilingual Named Entity Recognition Benchmark}}},
  shorttitle = {Universal {{NER}}},
  booktitle = {Proceedings of the 2024 {{Conference}} of the {{North American Chapter}} of the {{Association}} for {{Computational Linguistics}}: {{Human Language Technologies}} ({{Volume}} 1: {{Long Papers}})},
  author = {Mayhew, Stephen and Blevins, Terra and Liu, Shuheng and {\v S}uppa, Marek and Gonen, Hila and Imperial, Joseph Marvin and Karlsson, B{\"o}rje F. and Lin, Peiqin and Ljube{\v s}i{\'c}, Nikola and Miranda, {\relax LJ} and Plank, Barbara and Riabi, Arij and Pinter, Yuval},
  editor = {Duh, Kevin and Gomez, Helena and Bethard, Steven},
  year = {2024},
  month = jun,
  pages = {4322--4337},
  publisher = {Association for Computational Linguistics},
  address = {Mexico City, Mexico},
  doi = {10.18653/v1/2024.naacl-long.243},
  urldate = {2025-09-03}
}

@inproceedings{nedoluzhkoCorefUD10Coreference2022,
  title = {{{CorefUD}} 1.0: {{Coreference Meets Universal Dependencies}}},
  shorttitle = {{{CorefUD}} 1.0},
  booktitle = {Proceedings of the {{Thirteenth Language Resources}} and {{Evaluation Conference}}},
  author = {Nedoluzhko, Anna and Nov{\'a}k, Michal and Popel, Martin and {\v Z}abokrtsk{\'y}, Zden{\v e}k and Zeldes, Amir and Zeman, Daniel},
  editor = {Calzolari, Nicoletta and B{\'e}chet, Fr{\'e}d{\'e}ric and Blache, Philippe and Choukri, Khalid and Cieri, Christopher and Declerck, Thierry and Goggi, Sara and Isahara, Hitoshi and Maegaard, Bente and Mariani, Joseph and Mazo, H{\'e}l{\`e}ne and Odijk, Jan and Piperidis, Stelios},
  year = {2022},
  month = jun,
  pages = {4859--4872},
  publisher = {European Language Resources Association},
  address = {Marseille, France},
  urldate = {2025-09-03}
}

@inproceedings{hashimotoHighQualityMultilingualDataset2019,
  title = {A {{High-Quality Multilingual Dataset}} for {{Structured Documentation Translation}}},
  booktitle = {Proceedings of the {{Fourth Conference}} on {{Machine Translation}} ({{Volume}} 1: {{Research Papers}})},
  author = {Hashimoto, Kazuma and Buschiazzo, Raffaella and Bradbury, James and Marshall, Teresa and Socher, Richard and Xiong, Caiming},
  editor = {Bojar, Ond{\v r}ej and Chatterjee, Rajen and Federmann, Christian and Fishel, Mark and Graham, Yvette and Haddow, Barry and Huck, Matthias and Yepes, Antonio Jimeno and Koehn, Philipp and Martins, Andr{\'e} and Monz, Christof and Negri, Matteo and N{\'e}v{\'e}ol, Aur{\'e}lie and Neves, Mariana and Post, Matt and Turchi, Marco and Verspoor, Karin},
  year = {2019},
  month = aug,
  pages = {116--127},
  publisher = {Association for Computational Linguistics},
  address = {Florence, Italy},
  doi = {10.18653/v1/W19-5212},
  urldate = {2025-09-18}
}

@inproceedings{parekhContextualLabelProjection2024,
  title = {Contextual {{Label Projection}} for {{Cross-Lingual Structured Prediction}}},
  booktitle = {Proceedings of the 2024 {{Conference}} of the {{North American Chapter}} of the {{Association}} for {{Computational Linguistics}}: {{Human Language Technologies}} ({{Volume}} 1: {{Long Papers}})},
  author = {Parekh, Tanmay and Hsu, I-Hung and Huang, Kuan-Hao and Chang, Kai-Wei and Peng, Nanyun},
  year = {2024},
  pages = {5738--5757},
  publisher = {Association for Computational Linguistics},
  address = {Mexico City, Mexico},
  doi = {10.18653/v1/2024.naacl-long.321},
  urldate = {2025-09-18},
  langid = {english}
}

@inproceedings{artetxeCrosslingualTransferabilityMonolingual2020,
  title = {On the {{Cross-lingual Transferability}} of {{Monolingual Representations}}},
  booktitle = {Proceedings of the 58th {{Annual Meeting}} of the {{Association}} for {{Computational Linguistics}}},
  author = {Artetxe, Mikel and Ruder, Sebastian and Yogatama, Dani},
  editor = {Jurafsky, Dan and Chai, Joyce and Schluter, Natalie and Tetreault, Joel},
  year = {2020},
  month = jul,
  pages = {4623--4637},
  publisher = {Association for Computational Linguistics},
  address = {Online},
  doi = {10.18653/v1/2020.acl-main.421},
  urldate = {2025-09-18}
}

@inproceedings{lewisMLQAEvaluatingCrosslingual2020,
  title = {{{MLQA}}: {{Evaluating Cross-lingual Extractive Question Answering}}},
  shorttitle = {{{MLQA}}},
  booktitle = {Proceedings of the 58th {{Annual Meeting}} of the {{Association}} for {{Computational Linguistics}}},
  author = {Lewis, Patrick and Oguz, Barlas and Rinott, Ruty and Riedel, Sebastian and Schwenk, Holger},
  editor = {Jurafsky, Dan and Chai, Joyce and Schluter, Natalie and Tetreault, Joel},
  year = {2020},
  month = jul,
  pages = {7315--7330},
  publisher = {Association for Computational Linguistics},
  address = {Online},
  doi = {10.18653/v1/2020.acl-main.653},
  urldate = {2025-09-18}
}

@inproceedings{reiCOMET22UnbabelIST20222022,
  title = {{{COMET-22}}: {{Unbabel-IST}} 2022 {{Submission}} for the {{Metrics Shared Task}}},
  shorttitle = {{{COMET-22}}},
  booktitle = {Proceedings of the {{Seventh Conference}} on {{Machine Translation}} ({{WMT}})},
  author = {Rei, Ricardo and {C. de Souza}, Jos{\'e} G. and Alves, Duarte and Zerva, Chrysoula and Farinha, Ana C and Glushkova, Taisiya and Lavie, Alon and Coheur, Luisa and Martins, Andr{\'e} F. T.},
  editor = {Koehn, Philipp and Barrault, Lo{\"i}c and Bojar, Ond{\v r}ej and Bougares, Fethi and Chatterjee, Rajen and {Costa-juss{\`a}}, Marta R. and Federmann, Christian and Fishel, Mark and Fraser, Alexander and Freitag, Markus and Graham, Yvette and Grundkiewicz, Roman and Guzman, Paco and Haddow, Barry and Huck, Matthias and Jimeno Yepes, Antonio and Kocmi, Tom and Martins, Andr{\'e} and Morishita, Makoto and Monz, Christof and Nagata, Masaaki and Nakazawa, Toshiaki and Negri, Matteo and N{\'e}v{\'e}ol, Aur{\'e}lie and Neves, Mariana and Popel, Martin and Turchi, Marco and Zampieri, Marcos},
  year = {2022},
  month = dec,
  pages = {578--585},
  publisher = {Association for Computational Linguistics},
  address = {Abu Dhabi, United Arab Emirates (Hybrid)},
  urldate = {2025-09-22}
}

@misc{teamNoLanguageLeft2022,
  title = {No {{Language Left Behind}}: {{Scaling Human-Centered Machine Translation}}},
  shorttitle = {No {{Language Left Behind}}},
  author = {Team, {\relax NLLB} and {Costa-juss{\`a}}, Marta R. and Cross, James and {\c C}elebi, Onur and Elbayad, Maha and Heafield, Kenneth and Heffernan, Kevin and Kalbassi, Elahe and Lam, Janice and Licht, Daniel and Maillard, Jean and Sun, Anna and Wang, Skyler and Wenzek, Guillaume and Youngblood, Al and Akula, Bapi and Barrault, Loic and Gonzalez, Gabriel Mejia and Hansanti, Prangthip and Hoffman, John and Jarrett, Semarley and Sadagopan, Kaushik Ram and Rowe, Dirk and Spruit, Shannon and Tran, Chau and Andrews, Pierre and Ayan, Necip Fazil and Bhosale, Shruti and Edunov, Sergey and Fan, Angela and Gao, Cynthia and Goswami, Vedanuj and Guzm{\'a}n, Francisco and Koehn, Philipp and Mourachko, Alexandre and Ropers, Christophe and Saleem, Safiyyah and Schwenk, Holger and Wang, Jeff},
  year = {2022},
  month = aug,
  number = {arXiv:2207.04672},
  eprint = {2207.04672},
  primaryclass = {cs},
  publisher = {arXiv},
  doi = {10.48550/arXiv.2207.04672},
  urldate = {2025-09-22},
  archiveprefix = {arXiv}
}

@article{goyalFlores101EvaluationBenchmark2022,
  title = {The {{Flores-101 Evaluation Benchmark}} for {{Low-Resource}} and {{Multilingual Machine Translation}}},
  author = {Goyal, Naman and Gao, Cynthia and Chaudhary, Vishrav and Chen, Peng-Jen and Wenzek, Guillaume and Ju, Da and Krishnan, Sanjana and Ranzato, Marc'Aurelio and Guzm{\'a}n, Francisco and Fan, Angela},
  year = {2022},
  month = may,
  journal = {Transactions of the Association for Computational Linguistics},
  volume = {10},
  pages = {522--538},
  issn = {2307-387X},
  doi = {10.1162/tacl_a_00474},
  urldate = {2025-09-22}
}

@inproceedings{garcia-ferreroTProjectionHighQuality2023,
  title = {T-{{Projection}}: {{High Quality Annotation Projection}} for {{Sequence Labeling Tasks}}},
  shorttitle = {T-{{Projection}}},
  booktitle = {Findings of the {{Association}} for {{Computational Linguistics}}: {{EMNLP}} 2023},
  author = {{Garc{\'i}a-Ferrero}, Iker and Agerri, Rodrigo and Rigau, German},
  editor = {Bouamor, Houda and Pino, Juan and Bali, Kalika},
  year = {2023},
  month = dec,
  pages = {15203--15217},
  publisher = {Association for Computational Linguistics},
  address = {Singapore},
  doi = {10.18653/v1/2023.findings-emnlp.1015},
  urldate = {2025-09-24}
}

@inproceedings{douWordAlignmentFinetuning2021a,
  title = {Word {{Alignment}} by {{Fine-tuning Embeddings}} on {{Parallel Corpora}}},
  booktitle = {Proceedings of the 16th {{Conference}} of the {{European Chapter}} of the {{Association}} for {{Computational Linguistics}}: {{Main Volume}}},
  author = {Dou, Zi-Yi and Neubig, Graham},
  editor = {Merlo, Paola and Tiedemann, Jorg and Tsarfaty, Reut},
  year = {2021},
  month = apr,
  pages = {2112--2128},
  publisher = {Association for Computational Linguistics},
  address = {Online},
  doi = {10.18653/v1/2021.eacl-main.181},
  urldate = {2025-09-27}
}

@inproceedings{ebingDevilWordAlignment2025a,
  title = {The {{Devil Is}} in the {{Word Alignment Details}}: {{On Translation-Based Cross-Lingual Transfer}} for {{Token Classification Tasks}}},
  shorttitle = {The {{Devil Is}} in the {{Word Alignment Details}}},
  booktitle = {Findings of the {{Association}} for {{Computational Linguistics}}: {{ACL}} 2025},
  author = {Ebing, Benedikt and Glava{\v s}, Goran},
  editor = {Che, Wanxiang and Nabende, Joyce and Shutova, Ekaterina and Pilehvar, Mohammad Taher},
  year = {2025},
  month = jul,
  pages = {18111--18128},
  publisher = {Association for Computational Linguistics},
  address = {Vienna, Austria},
  doi = {10.18653/v1/2025.findings-acl.931},
  urldate = {2025-09-24},
  isbn = {979-8-89176-256-5}
}

@inproceedings{papineniBleuMethodAutomatic2002,
  title = {BLEU: A {{Method}} for {{Automatic Evaluation}} of {{Machine Translation}}},
  shorttitle = {Bleu},
  booktitle = {Proceedings of the 40th {{Annual Meeting}} of the {{Association}} for {{Computational Linguistics}}},
  author = {Papineni, Kishore and Roukos, Salim and Ward, Todd and Zhu, Wei-Jing},
  editor = {Isabelle, Pierre and Charniak, Eugene and Lin, Dekang},
  year = {2002},
  month = jul,
  pages = {311--318},
  publisher = {Association for Computational Linguistics},
  address = {Philadelphia, Pennsylvania, USA},
  doi = {10.3115/1073083.1073135},
  urldate = {2025-09-27}
}

@inproceedings{popovicChrFWordsHelping2017,
  title = {{{chrF}}++: Words Helping Character n-Grams},
  shorttitle = {{{chrF}}++},
  booktitle = {Proceedings of the {{Second Conference}} on {{Machine Translation}}},
  author = {Popovi{\'c}, Maja},
  editor = {Bojar, Ond{\v r}ej and Buck, Christian and Chatterjee, Rajen and Federmann, Christian and Graham, Yvette and Haddow, Barry and Huck, Matthias and Yepes, Antonio Jimeno and Koehn, Philipp and Kreutzer, Julia},
  year = {2017},
  month = sep,
  pages = {612--618},
  publisher = {Association for Computational Linguistics},
  address = {Copenhagen, Denmark},
  doi = {10.18653/v1/W17-4770},
  urldate = {2025-09-27}
}

@inproceedings{bakerCorpusLinguisticsTranslation1993,
  title = {Corpus {{Linguistics}} and {{Translation Studies}} --- {{Implications}} and {{Applications}}},
  booktitle = {Text and {{Technology}}},
  author = {Baker, Mona},
  editor = {Baker, Mona and Francis, Gill and {Tognini-Bonelli}, Elena},
  year = {1993},
  month = jun,
  pages = {233--250},
  publisher = {John Benjamins Publishing Company},
  address = {Amsterdam},
  doi = {10.1075/z.64.15bak},
  urldate = {2025-10-01},
  isbn = {978-90-272-2138-4 978-1-55619-494-8 978-90-272-8587-4},
  langid = {english}
}

@inproceedings{grahamTranslationeseMachineTranslation2019,
  title = {Statistical {{Power}} and {{Translationese}} in {{Machine Translation Evaluation}}},
  booktitle = {Proceedings of the 2020 {{Conference}} on {{Empirical Methods}} in {{Natural Language Processing}} ({{EMNLP}})},
  author = {Graham, Yvette and Haddow, Barry and Koehn, Philipp},
  editor = {Webber, Bonnie and Cohn, Trevor and He, Yulan and Liu, Yang},
  year = {2020},
  month = nov,
  pages = {72--81},
  publisher = {Association for Computational Linguistics},
  address = {Online},
  doi = {10.18653/v1/2020.emnlp-main.6},
  urldate = {2025-10-06}
}

@inproceedings{ebingTranslateNotTranslate2024,
  title = {To {{Translate}} or {{Not}} to {{Translate}}: {{A Systematic Investigation}} of {{Translation-Based Cross-Lingual Transfer}} to {{Low-Resource Languages}}},
  shorttitle = {To {{Translate}} or {{Not}} to {{Translate}}},
  booktitle = {Proceedings of the 2024 {{Conference}} of the {{North American Chapter}} of the {{Association}} for {{Computational Linguistics}}: {{Human Language Technologies}} ({{Volume}} 1: {{Long Papers}})},
  author = {Ebing, Benedikt and Glava{\v s}, Goran},
  editor = {Duh, Kevin and Gomez, Helena and Bethard, Steven},
  year = {2024},
  month = jun,
  pages = {5325--5344},
  publisher = {Association for Computational Linguistics},
  address = {Mexico City, Mexico},
  doi = {10.18653/v1/2024.naacl-long.298},
  urldate = {2025-09-24}
}

@inproceedings{huXTREMEMassivelyMultilingual2020,
  title = {{{XTREME}}: {{A Massively Multilingual Multi-task Benchmark}} for {{Evaluating Cross-lingual Generalisation}}},
  shorttitle = {{{XTREME}}},
  booktitle = {Proceedings of the 37th {{International Conference}} on {{Machine Learning}}},
  author = {Hu, Junjie and Ruder, Sebastian and Siddhant, Aditya and Neubig, Graham and Firat, Orhan and Johnson, Melvin},
  year = {2020},
  month = nov,
  pages = {4411--4421},
  publisher = {PMLR},
  issn = {2640-3498},
  urldate = {2025-10-01},
  langid = {english}
}

@inproceedings{hannemanHowShouldMarkup2020,
  title = {How {{Should Markup Tags Be Translated}}?},
  booktitle = {Proceedings of the {{Fifth Conference}} on {{Machine Translation}}},
  author = {Hanneman, Greg and Dinu, Georgiana},
  editor = {Barrault, Lo{\"i}c and Bojar, Ond{\v r}ej and Bougares, Fethi and Chatterjee, Rajen and {Costa-juss{\`a}}, Marta R. and Federmann, Christian and Fishel, Mark and Fraser, Alexander and Graham, Yvette and Guzman, Paco and Haddow, Barry and Huck, Matthias and Yepes, Antonio Jimeno and Koehn, Philipp and Martins, Andr{\'e} and Morishita, Makoto and Monz, Christof and Nagata, Masaaki and Nakazawa, Toshiaki and Negri, Matteo},
  year = {2020},
  month = nov,
  pages = {1160--1173},
  publisher = {Association for Computational Linguistics},
  address = {Online},
  urldate = {2025-10-01}
}

@inproceedings{joanisTransferringMarkupTags2013,
  title = {Transferring Markup Tags in Statistical Machine Translation: A Two-Stream Approach},
  shorttitle = {Transferring Markup Tags in Statistical Machine Translation},
  booktitle = {Proceedings of the 2nd {{Workshop}} on {{Post-editing Technology}} and {{Practice}}},
  author = {Joanis, Eric and Stewart, Darlene and Larkin, Samuel and Kuhn, Roland},
  editor = {O'Brien, Sharon and Simard, Michel and Specia, Lucia},
  year = {2013},
  month = sep,
  address = {Nice, France},
  urldate = {2025-10-01}
}

@inproceedings{mullerTreatmentMarkupStatistical2017,
  title = {Treatment of {{Markup}} in {{Statistical Machine Translation}}},
  booktitle = {Proceedings of the {{Third Workshop}} on {{Discourse}} in {{Machine Translation}}},
  author = {M{\"u}ller, Mathias},
  editor = {Webber, Bonnie and {Popescu-Belis}, Andrei and Tiedemann, J{\"o}rg},
  year = {2017},
  month = sep,
  pages = {36--46},
  publisher = {Association for Computational Linguistics},
  address = {Copenhagen, Denmark},
  doi = {10.18653/v1/W17-4804},
  urldate = {2025-10-01}
}

@inproceedings{bammanTransferringStructuralMarkup2010,
  title = {Transferring Structural Markup across Translations Using Multilingual Alignment and Projection},
  booktitle = {Proceedings of the 10th Annual Joint Conference on {{Digital}} Libraries},
  author = {Bamman, David and Babeu, Alison and Crane, Gregory},
  year = {2010},
  month = jun,
  series = {{{JCDL}} '10},
  pages = {11--20},
  publisher = {Association for Computing Machinery},
  address = {New York, NY, USA},
  doi = {10.1145/1816123.1816126},
  urldate = {2025-10-01},
  isbn = {978-1-4503-0085-8}
}

@inproceedings{buschbeckMultilingualMultiwayEvaluation2022,
  title = {A {{Multilingual Multiway Evaluation Data Set}} for {{Structured Document Translation}} of {{Asian Languages}}},
  booktitle = {Findings of the {{Association}} for {{Computational Linguistics}}: {{AACL-IJCNLP}} 2022},
  author = {Buschbeck, Bianka and Dabre, Raj and Exel, Miriam and Huck, Matthias and Huy, Patrick and Rubino, Raphael and Tanaka, Hideki},
  year = {2022},
  pages = {237--245},
  publisher = {Association for Computational Linguistics},
  address = {Online only},
  doi = {10.18653/v1/2022.findings-aacl.23},
  urldate = {2025-10-01},
  langid = {english}
}

@inproceedings{dabreNICTsSubmissionWAT2022,
  title = {{{NICT}}'s {{Submission}} to the {{WAT}} 2022 {{Structured Document Translation Task}}},
  booktitle = {Proceedings of the 9th {{Workshop}} on {{Asian Translation}}},
  author = {Dabre, Raj},
  year = {2022},
  month = oct,
  pages = {64--67},
  publisher = {International Conference on Computational Linguistics},
  address = {Gyeongju, Republic of Korea},
  urldate = {2025-10-01}
}

@inproceedings{dabreStudyEffectivenessLarge2023,
  title = {A {{Study}} on the {{Effectiveness}} of {{Large Language Models}} for {{Translation}} with {{Markup}}},
  booktitle = {Proceedings of {{Machine Translation Summit XIX}}, {{Vol}}. 1: {{Research Track}}},
  author = {Dabre, Raj and Buschbeck, Bianka and Exel, Miriam and Tanaka, Hideki},
  editor = {Utiyama, Masao and Wang, Rui},
  year = {2023},
  month = sep,
  pages = {148--159},
  publisher = {Asia-Pacific Association for Machine Translation},
  address = {Macau SAR, China},
  urldate = {2025-10-01}
}

@inproceedings{conneauUnsupervisedCrosslingualRepresentation2020,
  title = {Unsupervised {{Cross-lingual Representation Learning}} at {{Scale}}},
  booktitle = {Proceedings of the 58th {{Annual Meeting}} of the {{Association}} for {{Computational Linguistics}}},
  author = {Conneau, Alexis and Khandelwal, Kartikay and Goyal, Naman and Chaudhary, Vishrav and Wenzek, Guillaume and Guzm{\'a}n, Francisco and Grave, Edouard and Ott, Myle and Zettlemoyer, Luke and Stoyanov, Veselin},
  editor = {Jurafsky, Dan and Chai, Joyce and Schluter, Natalie and Tetreault, Joel},
  year = {2020},
  month = jul,
  pages = {8440--8451},
  publisher = {Association for Computational Linguistics},
  address = {Online},
  doi = {10.18653/v1/2020.acl-main.747},
  urldate = {2025-10-02}
}

@misc{corefud13,
             title = {Coreference in Universal Dependencies 1.3 ({CorefUD} 1.3)},
 author = {Nov{\'a}k, Michal and Popel, Martin and Zeman, Daniel and {\v Z}abokrtsk{\'y}, Zden{\v e}k and Nedoluzhko, Anna and Acar, Kutay and Bamman, David and Bourgonje, Peter and Cinkov{\'a}, Silvie and Eckhoff, Hanne and Cebiro{\u g}lu Eryi{\u g}it, G{\"u}l{\c s}en and Haji{\v c}, Jan and Hardmeier, Christian and Haug, Dag and J{\o}rgensen, Tollef and K{\aa}sen, Andre and Krielke, Pauline and Landragin, Fr{\'e}d{\'e}ric and Lapshinova-Koltunski, Ekaterina and M{\ae}hlum, Petter and Mart{\'{\i}}, M. Ant{\`o}nia and Mikulov{\'a}, Marie and Milintsevich, Kirill and Mujadia, Vandan and Muzerelle, Judith and Nam, Sangha and N{\o}klestad, Anders and Ogrodniczuk, Maciej and {\O}vrelid, Lilja and Pamay Arslan, Tu{\u g}ba and Porada, Ian and Recasens, Marta and Solberg, Per Erik and Stede, Manfred and Straka, Milan and Swanson, Daniel and Toldova, Svetlana and Vad{\'a}sz, No{\'e}mi and Velldal, Erik and Vincze, Veronika and Zeldes, Amir and {\v Z}itkus, Voldemaras},
 url = {http://hdl.handle.net/11234/1-5896},
 note = {{LINDAT}/{CLARIAH}-{CZ} digital library at the Institute of Formal and Applied Linguistics ({{\'U}FAL})},
 copyright = {Licence {CorefUD} v1.3},
 year = {2025}
}

@inproceedings{novakFindingsThirdShared2024,
  title = {Findings of the {{Third Shared Task}} on {{Multilingual Coreference Resolution}}},
  booktitle = {Proceedings of the {{Seventh Workshop}} on {{Computational Models}} of {{Reference}}, {{Anaphora}} and {{Coreference}}},
  author = {Nov{\'a}k, Michal and Dohnalov{\'a}, Barbora and Konopik, Miloslav and Nedoluzhko, Anna and Popel, Martin and Prazak, Ondrej and Sido, Jakub and Straka, Milan and {\v Z}abokrtsk{\'y}, Zden{\v e}k and Zeman, Daniel},
  editor = {Ogrodniczuk, Maciej and Nedoluzhko, Anna and Poesio, Massimo and Pradhan, Sameer and Ng, Vincent},
  year = {2024},
  month = nov,
  pages = {78--96},
  publisher = {Association for Computational Linguistics},
  address = {Miami},
  doi = {10.18653/v1/2024.crac-1.8},
  urldate = {2025-10-02}
}

@inproceedings{prazakMultilingualCoreferenceResolution2021,
  title = {Multilingual {{Coreference Resolution}} with {{Harmonized Annotations}}},
  booktitle = {Proceedings of the {{International Conference}} on {{Recent Advances}} in {{Natural Language Processing}} ({{RANLP}} 2021)},
  author = {Pra{\v z}{\'a}k, Ond{\v r}ej and Konop{\'i}k, Miloslav and Sido, Jakub},
  editor = {Mitkov, Ruslan and Angelova, Galia},
  year = {2021},
  month = sep,
  pages = {1119--1123},
  publisher = {INCOMA Ltd.},
  address = {Held Online},
  urldate = {2025-10-02}
}

@inproceedings{devlinBERTPretrainingDeep2019,
  title = {{{BERT}}: {{Pre-training}} of {{Deep Bidirectional Transformers}} for {{Language Understanding}}},
  shorttitle = {{{BERT}}},
  booktitle = {Proceedings of the 2019 {{Conference}} of the {{North American Chapter}} of the {{Association}} for {{Computational Linguistics}}: {{Human Language Technologies}}, {{Volume}} 1 ({{Long}} and {{Short Papers}})},
  author = {Devlin, Jacob and Chang, Ming-Wei and Lee, Kenton and Toutanova, Kristina},
  editor = {Burstein, Jill and Doran, Christy and Solorio, Thamar},
  year = {2019},
  month = jun,
  pages = {4171--4186},
  publisher = {Association for Computational Linguistics},
  address = {Minneapolis, Minnesota},
  doi = {10.18653/v1/N19-1423},
  urldate = {2025-10-02}
}

@misc{weischedelralphOntoNotesRelease502013,
  title = {{{OntoNotes Release}} 5.0},
  author = {{Weischedel, Ralph} and {Palmer, Martha} and {Marcus, Mitchell} and {Hovy, Eduard} and {Pradhan, Sameer} and {Ramshaw, Lance} and {Xue, Nianwen} and {Taylor, Ann} and {Kaufman, Jeff} and {Franchini, Michelle} and {El-Bachouti, Mohammed} and {Belvin, Robert} and {Houston, Ann}},
  year = {2013},
  month = oct,
  pages = {2806280 KB},
  publisher = {Linguistic Data Consortium},
  note = {Linguistic Data Consortium},
  doi = {10.35111/XMHB-2B84},
  urldate = {2025-10-02}
}

@inproceedings{rajpurkarSQuAD100000Questions2016,
  title = {{{SQuAD}}: 100,000+ {{Questions}} for {{Machine Comprehension}} of {{Text}}},
  shorttitle = {{{SQuAD}}},
  booktitle = {Proceedings of the 2016 {{Conference}} on {{Empirical Methods}} in {{Natural Language Processing}}},
  author = {Rajpurkar, Pranav and Zhang, Jian and Lopyrev, Konstantin and Liang, Percy},
  editor = {Su, Jian and Duh, Kevin and Carreras, Xavier},
  year = {2016},
  month = nov,
  pages = {2383--2392},
  publisher = {Association for Computational Linguistics},
  address = {Austin, Texas},
  doi = {10.18653/v1/D16-1264},
  urldate = {2025-10-02}
}

@misc{bucherFineTunedSmallLLMs2024,
  title = {Fine-{{Tuned}} '{{Small}}' {{LLMs}} ({{Still}}) {{Significantly Outperform Zero-Shot Generative AI Models}} in {{Text Classification}}},
  author = {Bucher, Martin Juan Jos{\'e} and Martini, Marco},
  year = {2024},
  month = aug,
  number = {arXiv:2406.08660},
  eprint = {2406.08660},
  primaryclass = {cs},
  publisher = {arXiv},
  doi = {10.48550/arXiv.2406.08660},
  urldate = {2025-10-03},
  archiveprefix = {arXiv}
}

@article{luLargeLanguageModels2025,
  title = {Large {{Language Models Struggle}} in {{Token-Level Clinical Named Entity Recognition}}},
  author = {Lu, Qiuhao and Li, Rui and Wen, Andrew and Wang, Jinlian and Wang, Liwei and Liu, Hongfang},
  year = {2025},
  month = may,
  journal = {AMIA Annual Symposium Proceedings},
  volume = {2024},
  pages = {748--757},
  issn = {1942-597X},
  urldate = {2025-10-03},
  pmcid = {PMC12099373},
  pmid = {40417588}
}

@inproceedings{poradaControlledReevaluationCoreference2024a,
  title = {A {{Controlled Reevaluation}} of {{Coreference Resolution Models}}},
  booktitle = {Proceedings of the 2024 {{Joint International Conference}} on {{Computational Linguistics}}, {{Language Resources}} and {{Evaluation}} ({{LREC-COLING}} 2024)},
  author = {Porada, Ian and Zou, Xiyuan and Cheung, Jackie Chi Kit},
  editor = {Calzolari, Nicoletta and Kan, Min-Yen and Hoste, Veronique and Lenci, Alessandro and Sakti, Sakriani and Xue, Nianwen},
  year = {2024},
  month = may,
  pages = {256--263},
  publisher = {{ELRA and ICCL}},
  address = {Torino, Italia},
  urldate = {2025-10-03}
}

@inproceedings{weiAreLLMsGood2024,
  title = {Are {{LLMs Good Annotators}} for {{Discourse-level Event Relation Extraction}}?},
  booktitle = {Findings of the {{Association}} for {{Computational Linguistics}}: {{EMNLP}} 2024},
  author = {Wei, Kangda and Gautam, Aayush and Huang, Ruihong},
  editor = {{Al-Onaizan}, Yaser and Bansal, Mohit and Chen, Yun-Nung},
  year = {2024},
  month = nov,
  pages = {1--19},
  publisher = {Association for Computational Linguistics},
  address = {Miami, Florida, USA},
  doi = {10.18653/v1/2024.findings-emnlp.1},
  urldate = {2025-10-03}
}

@inproceedings{liuStructuredSpanSelector2022,
  title = {A {{Structured Span Selector}}},
  booktitle = {Proceedings of the 2022 {{Conference}} of the {{North American Chapter}} of the {{Association}} for {{Computational Linguistics}}: {{Human Language Technologies}}},
  author = {Liu, Tianyu and Jiang, Yuchen and Cotterell, Ryan and Sachan, Mrinmaya},
  editor = {Carpuat, Marine and {de Marneffe}, Marie-Catherine and Meza Ruiz, Ivan Vladimir},
  year = {2022},
  month = jul,
  pages = {2629--2641},
  publisher = {Association for Computational Linguistics},
  address = {Seattle, WA, USA},
  doi = {10.18653/v1/2022.naacl-main.189},
  urldate = {2025-10-03}
}

@article{black2004ratcliff,
  title={Ratcliff/Obershelp pattern recognition},
  author={Black, Paul E},
  journal={Dictionary of algorithms and data structures},
  volume={17},
  year={2004},
  publisher={National Institute of Standards and Technology Gaithersburg, MD, USA}
}

@inproceedings{dazaXSRLParallelCrossLingual2020a,
  title = {X-{{SRL}}: {{A Parallel Cross-Lingual Semantic Role Labeling Dataset}}},
  shorttitle = {X-{{SRL}}},
  booktitle = {Proceedings of the 2020 {{Conference}} on {{Empirical Methods}} in {{Natural Language Processing}} ({{EMNLP}})},
  author = {Daza, Angel and Frank, Anette},
  editor = {Webber, Bonnie and Cohn, Trevor and He, Yulan and Liu, Yang},
  year = {2020},
  month = nov,
  pages = {3904--3914},
  publisher = {Association for Computational Linguistics},
  address = {Online},
  doi = {10.18653/v1/2020.emnlp-main.321},
  urldate = {2025-10-05}
}

@inproceedings{moradshahiLocalizingOpenOntologyQA2020,
  title = {Localizing {{Open-Ontology QA Semantic Parsers}} in a {{Day Using Machine Translation}}},
  booktitle = {Proceedings of the 2020 {{Conference}} on {{Empirical Methods}} in {{Natural Language Processing}} ({{EMNLP}})},
  author = {Moradshahi, Mehrad and Campagna, Giovanni and Semnani, Sina and Xu, Silei and Lam, Monica},
  editor = {Webber, Bonnie and Cohn, Trevor and He, Yulan and Liu, Yang},
  year = {2020},
  month = nov,
  pages = {5970--5983},
  publisher = {Association for Computational Linguistics},
  address = {Online},
  doi = {10.18653/v1/2020.emnlp-main.481},
  urldate = {2025-10-06}
}

@inproceedings{niWeaklySupervisedCrossLingual2017,
  title = {Weakly {{Supervised Cross-Lingual Named Entity Recognition}} via {{Effective Annotation}} and {{Representation Projection}}},
  booktitle = {Proceedings of the 55th {{Annual Meeting}} of the {{Association}} for {{Computational Linguistics}} ({{Volume}} 1: {{Long Papers}})},
  author = {Ni, Jian and Dinu, Georgiana and Florian, Radu},
  editor = {Barzilay, Regina and Kan, Min-Yen},
  year = {2017},
  month = jul,
  pages = {1470--1480},
  publisher = {Association for Computational Linguistics},
  address = {Vancouver, Canada},
  doi = {10.18653/v1/P17-1135},
  urldate = {2025-10-06}
}

@inproceedings{louTranslationBasedImplicitAnnotation2022,
  title = {Translation-{{Based Implicit Annotation Projection}} for {{Zero-Shot Cross-Lingual Event Argument Extraction}}},
  booktitle = {Proceedings of the 45th {{International ACM SIGIR Conference}} on {{Research}} and {{Development}} in {{Information Retrieval}}},
  author = {Lou, Chenwei and Gao, Jun and Yu, Changlong and Wang, Wei and Zhao, Huan and Tu, Weiwei and Xu, Ruifeng},
  year = {2022},
  month = jul,
  series = {{{SIGIR}} '22},
  pages = {2076--2081},
  publisher = {Association for Computing Machinery},
  address = {New York, NY, USA},
  doi = {10.1145/3477495.3531808},
  urldate = {2025-10-05},
  isbn = {978-1-4503-8732-3}
}

@inproceedings{chenFrustratinglyEasyLabel2024,
  title = {Frustratingly {{Easy Label Projection}} for {{Cross-lingual Transfer}}},
  booktitle = {Findings of the {{Association}} for {{Computational Linguistics}}: {{ACL}} 2023},
  author = {Chen, Yang and Jiang, Chao and Ritter, Alan and Xu, Wei},
  editor = {Rogers, Anna and {Boyd-Graber}, Jordan and Okazaki, Naoaki},
  year = {2023},
  month = jul,
  pages = {5775--5796},
  publisher = {Association for Computational Linguistics},
  address = {Toronto, Canada},
  doi = {10.18653/v1/2023.findings-acl.357},
  urldate = {2025-10-06}
}

@inproceedings{leConstrainedDecodingCrosslingual2024,
title={Constrained Decoding for Cross-lingual Label Projection},
author={Duong Minh Le and Yang Chen and Alan Ritter and Wei Xu},
booktitle={The Twelfth International Conference on Learning Representations},
year={2024},
address={Vienna, Austria},
url={https://openreview.net/forum?id=DayPQKXaQk}
}

@misc{teamGemma3Technical2025,
  title = {Gemma 3 {{Technical Report}}},
  author = {Team, Gemma and Kamath, Aishwarya and Ferret, Johan and Pathak, Shreya and Vieillard, Nino and Merhej, Ramona and Perrin, Sarah and Matejovicova, Tatiana and Ram{\'e}, Alexandre and Rivi{\`e}re, Morgane and Rouillard, Louis and Mesnard, Thomas and Cideron, Geoffrey and Grill, Jean-bastien and Ramos, Sabela and Yvinec, Edouard and Casbon, Michelle and Pot, Etienne and Penchev, Ivo and Liu, Ga{\"e}l and Visin, Francesco and Kenealy, Kathleen and Beyer, Lucas and Zhai, Xiaohai and Tsitsulin, Anton and {Busa-Fekete}, Robert and Feng, Alex and Sachdeva, Noveen and Coleman, Benjamin and Gao, Yi and Mustafa, Basil and Barr, Iain and Parisotto, Emilio and Tian, David and Eyal, Matan and Cherry, Colin and Peter, Jan-Thorsten and Sinopalnikov, Danila and Bhupatiraju, Surya and Agarwal, Rishabh and Kazemi, Mehran and Malkin, Dan and Kumar, Ravin and Vilar, David and Brusilovsky, Idan and Luo, Jiaming and Steiner, Andreas and Friesen, Abe and Sharma, Abhanshu and Sharma, Abheesht and Gilady, Adi Mayrav and Goedeckemeyer, Adrian and Saade, Alaa and Feng, Alex and Kolesnikov, Alexander and Bendebury, Alexei and Abdagic, Alvin and Vadi, Amit and Gy{\"o}rgy, Andr{\'a}s and Pinto, Andr{\'e} Susano and Das, Anil and Bapna, Ankur and Miech, Antoine and Yang, Antoine and Paterson, Antonia and Shenoy, Ashish and Chakrabarti, Ayan and Piot, Bilal and Wu, Bo and Shahriari, Bobak and Petrini, Bryce and Chen, Charlie and Lan, Charline Le and {Choquette-Choo}, Christopher A. and Carey, C. J. and Brick, Cormac and Deutsch, Daniel and Eisenbud, Danielle and Cattle, Dee and Cheng, Derek and Paparas, Dimitris and Sreepathihalli, Divyashree Shivakumar and Reid, Doug and Tran, Dustin and Zelle, Dustin and Noland, Eric and Huizenga, Erwin and Kharitonov, Eugene and Liu, Frederick and Amirkhanyan, Gagik and Cameron, Glenn and Hashemi, Hadi and {Klimczak-Pluci{\'n}ska}, Hanna and Singh, Harman and Mehta, Harsh and Lehri, Harshal Tushar and Hazimeh, Hussein and Ballantyne, Ian and Szpektor, Idan and Nardini, Ivan and {Pouget-Abadie}, Jean and Chan, Jetha and Stanton, Joe and Wieting, John and Lai, Jonathan and Orbay, Jordi and Fernandez, Joseph and Newlan, Josh and Ji, Ju-yeong and Singh, Jyotinder and Black, Kat and Yu, Kathy and Hui, Kevin and Vodrahalli, Kiran and Greff, Klaus and Qiu, Linhai and Valentine, Marcella and Coelho, Marina and Ritter, Marvin and Hoffman, Matt and Watson, Matthew and Chaturvedi, Mayank and Moynihan, Michael and Ma, Min and Babar, Nabila and Noy, Natasha and Byrd, Nathan and Roy, Nick and Momchev, Nikola and Chauhan, Nilay and Sachdeva, Noveen and Bunyan, Oskar and Botarda, Pankil and Caron, Paul and Rubenstein, Paul Kishan and Culliton, Phil and Schmid, Philipp and Sessa, Pier Giuseppe and Xu, Pingmei and Stanczyk, Piotr and Tafti, Pouya and Shivanna, Rakesh and Wu, Renjie and Pan, Renke and Rokni, Reza and Willoughby, Rob and Vallu, Rohith and Mullins, Ryan and Jerome, Sammy and Smoot, Sara and Girgin, Sertan and Iqbal, Shariq and Reddy, Shashir and Sheth, Shruti and P{\~o}der, Siim and Bhatnagar, Sijal and Panyam, Sindhu Raghuram and Eiger, Sivan and Zhang, Susan and Liu, Tianqi and Yacovone, Trevor and Liechty, Tyler and Kalra, Uday and Evci, Utku and Misra, Vedant and Roseberry, Vincent and Feinberg, Vlad and Kolesnikov, Vlad and Han, Woohyun and Kwon, Woosuk and Chen, Xi and Chow, Yinlam and Zhu, Yuvein and Wei, Zichuan and Egyed, Zoltan and Cotruta, Victor and Giang, Minh and Kirk, Phoebe and Rao, Anand and Black, Kat and Babar, Nabila and Lo, Jessica and Moreira, Erica and Martins, Luiz Gustavo and Sanseviero, Omar and Gonzalez, Lucas and Gleicher, Zach and Warkentin, Tris and Mirrokni, Vahab and Senter, Evan and Collins, Eli and Barral, Joelle and Ghahramani, Zoubin and Hadsell, Raia and Matias, Yossi and Sculley, D. and Petrov, Slav and Fiedel, Noah and Shazeer, Noam and Vinyals, Oriol and Dean, Jeff and Hassabis, Demis and Kavukcuoglu, Koray and Farabet, Clement and Buchatskaya, Elena and Alayrac, Jean-Baptiste and Anil, Rohan and Dmitry and Lepikhin and Borgeaud, Sebastian and Bachem, Olivier and Joulin, Armand and Andreev, Alek and Hardin, Cassidy and Dadashi, Robert and Hussenot, L{\'e}onard},
  year = {2025},
  month = mar,
  number = {arXiv:2503.19786},
  eprint = {2503.19786},
  primaryclass = {cs},
  publisher = {arXiv},
  doi = {10.48550/arXiv.2503.19786},
  urldate = {2025-12-20},
  archiveprefix = {arXiv}
}

@inproceedings{liuConditionsCatastrophicForgetting2025,
  title = {Conditions for {{Catastrophic Forgetting}} in {{Multilingual Translation}}},
  booktitle = {Proceedings of the 5th {{Workshop}} on {{Multilingual Representation Learning}} ({{MRL}} 2025)},
  author = {Liu, Danni and Niehues, Jan},
  editor = {Adelani, David Ifeoluwa and Arnett, Catherine and Ataman, Duygu and Chang, Tyler A. and Gonen, Hila and Raja, Rahul and Schmidt, Fabian and Stap, David and Wang, Jiayi},
  year = {2025},
  month = nov,
  pages = {347--359},
  publisher = {Association for Computational Linguistics},
  address = {Suzhuo, China},
  doi = {10.18653/v1/2025.mrl-main.23},
  urldate = {2025-12-24},
  isbn = {979-8-89176-345-6}
}

@inproceedings{akbikGeneratingHighQuality2015,
  title = {Generating {{High Quality Proposition Banks}} for {{Multilingual Semantic Role Labeling}}},
  booktitle = {Proceedings of the 53rd {{Annual Meeting}} of the {{Association}} for {{Computational Linguistics}} and the 7th {{International Joint Conference}} on {{Natural Language Processing}} ({{Volume}} 1: {{Long Papers}})},
  author = {Akbik, Alan and Chiticariu, Laura and Danilevsky, Marina and Li, Yunyao and Vaithyanathan, Shivakumar and Zhu, Huaiyu},
  editor = {Zong, Chengqing and Strube, Michael},
  year = {2015},
  month = jul,
  pages = {397--407},
  publisher = {Association for Computational Linguistics},
  address = {Beijing, China},
  doi = {10.3115/v1/P15-1039},
  urldate = {2025-12-27}
}

@inproceedings{aminianTransferringSemanticRoles2017,
  title = {Transferring {{Semantic Roles Using Translation}} and {{Syntactic Information}}},
  booktitle = {Proceedings of the {{Eighth International Joint Conference}} on {{Natural Language Processing}} ({{Volume}} 2: {{Short Papers}})},
  author = {Aminian, Maryam and Rasooli, Mohammad Sadegh and Diab, Mona},
  editor = {Kondrak, Greg and Watanabe, Taro},
  year = {2017},
  month = nov,
  pages = {13--19},
  publisher = {Asian Federation of Natural Language Processing},
  address = {Taipei, Taiwan},
  urldate = {2025-12-27}
}

\clearpage
\appendix

\section{Training}
\label{sec:appendix:training}

\begin{table}[h]
\centering
\small
\begin{tabular}{l l}
\toprule
\textbf{Setting} & \textbf{Value} \\
\midrule
Learning rate & 1e\textsuperscript{-3} \\
Batch size & 8 \\
Grad. Accumulation & 2 \\
Scheduler & Inverse square root \\
Weight Decay & 0.01 \\
Warmup & 5\% steps \\
Precision & \texttt{bfloat16} \\
\bottomrule
\end{tabular}
\caption{Relevant hyperparameters for LabelPigeon fine-tuning.}
\label{tab:train-hparams}
\end{table}

\begin{table*}[ht]
\centering
\begin{tabular}{lcccccc}
\toprule
 & \multicolumn{2}{c}{\texttt{de-en}} & \multicolumn{2}{c}{\texttt{ru-en}} & \multicolumn{2}{c}{\texttt{zh-en}} \\
\cmidrule(lr){2-3} \cmidrule(lr){4-5} \cmidrule(lr){6-7}
 & \textbf{Train} & \textbf{Valid} & \textbf{Train} & \textbf{Valid} & \textbf{Train} & \textbf{Valid} \\
\midrule
\textbf{Samples (N)} & 24311 & 1262 & 24243 & 1301 & 24173 & 1248 \\
\textbf{Total Tags} & 41569 & 2179 & 41542 & 2250 & 41881 & 2122 \\
\textbf{Max Tags / Example} & 50 & 9 & 50 & 14 & 50 & 13 \\
\textbf{Max Unique Tags / Example} & 6 & 5 & 6 & 5 & 6 & 5 \\
\textbf{Avg. \# Tags / Example} & 1.71 & 1.73 & 1.71 & 1.73 & 1.73 & 1.70 \\
\bottomrule
\end{tabular}
\caption{Statistics for our training data.}
\label{tab:training-data-stats}
\end{table*}

As described in \S \ref{sec:labelpigeon}, we use the Salesforce Localization XML MT dataset provided by \citet{hashimotoHighQualityMultilingualDataset2019}, modified for label projection. Relevant statistics after filtering are compiled in Table \ref{tab:training-data-stats}. The model is trained with the hyperparameters given in Table \ref{tab:train-hparams}. We note that since the original dataset includes examples with multiple instances of the same tag, the total number of tags is higher than the unique number of tags. As we filter out instances without tags, the minimum number of tags is 1 for all training data subsets.

\subsection{Ablations}
\label{sec:appendix:ablations}

\begin{table*}[t]
  \centering
  \begin{tabular}{l r r r r r r r r}
  \toprule
  \multirow[c]{2}{*}{\textbf{Language}} & \multicolumn{4}{c}{\textbf{COMET Score}} & \multicolumn{4}{c}{\textbf{Label Matches (F1, \%)}} \\
  \cmidrule(lr){2-5} \cmidrule(lr){6-9}
   & \textbf{Base} & \textbf{One} & \textbf{Some} & \textbf{All} & \textbf{Base} & \textbf{One} & \textbf{Some} & \textbf{All} \\
  \midrule
  \multicolumn{9}{l}{\textbf{XQUAD}} \\
  \cmidrule(lr){1-9}
  Arabic & 79.9 & \cellcolor{ForestGreen!4}80.5 & \cellcolor{ForestGreen!12}\textbf{\textcolor{black!85}{81.4}} & \cellcolor{ForestGreen!2}80.2 & 2.3 & \cellcolor{ForestGreen!35}71.9 & \cellcolor{ForestGreen!37}\textbf{\textcolor{black!85}{75.7}} & \cellcolor{ForestGreen!36}74.7 \\
  Chinese & 79.8 & \cellcolor{BrickRed!5}79.2 & \cellcolor{ForestGreen!4}\textbf{\textcolor{black!85}{80.3}} & \cellcolor{ForestGreen!4}80.3 & 4.5 & \cellcolor{ForestGreen!33}70.3 & \cellcolor{ForestGreen!36}\textbf{\textcolor{black!85}{76.5}} & \cellcolor{ForestGreen!35}75.0 \\
  German & 81.6 & \cellcolor{ForestGreen!6}82.4 & \cellcolor{ForestGreen!13}\textbf{\textcolor{black!85}{83.2}} & \cellcolor{ForestGreen!4}82.2 & 6.7 & \cellcolor{ForestGreen!39}83.9 & \cellcolor{ForestGreen!40}\textbf{\textcolor{black!85}{86.2}} & \cellcolor{ForestGreen!39}84.8 \\
  Greek & 82.6 & \cellcolor{ForestGreen!4}83.1 & \cellcolor{ForestGreen!15}\textbf{\textcolor{black!85}{84.5}} & \cellcolor{ForestGreen!4}83.1 & 3.4 & \cellcolor{ForestGreen!35}73.8 & \cellcolor{ForestGreen!37}\textbf{\textcolor{black!85}{76.6}} & \cellcolor{ForestGreen!36}75.4 \\
  Hindi & 80.7 & \cellcolor{ForestGreen!1}80.8 & \cellcolor{ForestGreen!4}\textbf{\textcolor{black!85}{81.2}} & \cellcolor{BrickRed!0}80.7 & 7.7 & \cellcolor{ForestGreen!35}78.4 & \cellcolor{ForestGreen!36}\textbf{\textcolor{black!85}{80.4}} & \cellcolor{ForestGreen!36}79.8 \\
  Romanian & 82.7 & \cellcolor{ForestGreen!7}83.6 & \cellcolor{ForestGreen!14}\textbf{\textcolor{black!85}{84.5}} & \cellcolor{ForestGreen!3}83.1 & 8.9 & \cellcolor{ForestGreen!37}82.9 & \cellcolor{ForestGreen!38}\textbf{\textcolor{black!85}{85.2}} & \cellcolor{ForestGreen!38}84.3 \\
  Russian & 81.2 & \cellcolor{ForestGreen!10}82.5 & \cellcolor{ForestGreen!17}\textbf{\textcolor{black!85}{83.3}} & \cellcolor{ForestGreen!9}82.3 & 7.6 & \cellcolor{ForestGreen!35}77.6 & \cellcolor{ForestGreen!36}\textbf{\textcolor{black!85}{79.8}} & \cellcolor{ForestGreen!36}79.4 \\
  Spanish & 83.4 & \cellcolor{ForestGreen!6}84.1 & \cellcolor{ForestGreen!9}\textbf{\textcolor{black!85}{84.5}} & \cellcolor{ForestGreen!3}83.8 & 3.6 & \cellcolor{ForestGreen!41}86.0 & \cellcolor{ForestGreen!43}\textbf{\textcolor{black!85}{88.6}} & \cellcolor{ForestGreen!42}88.3 \\
  Thai & 78.3 & \cellcolor{BrickRed!3}78.0 & \cellcolor{ForestGreen!1}\textbf{\textcolor{black!85}{78.4}} & \cellcolor{BrickRed!7}77.5 & 7.9 & \cellcolor{ForestGreen!28}64.5 & \cellcolor{ForestGreen!30}\textbf{\textcolor{black!85}{67.0}} & \cellcolor{ForestGreen!29}65.9 \\
  Turkish & 82.4 & \cellcolor{ForestGreen!14}84.2 & \cellcolor{ForestGreen!20}\textbf{\textcolor{black!85}{84.9}} & \cellcolor{ForestGreen!11}83.8 & 7.5 & \cellcolor{ForestGreen!36}79.2 & \cellcolor{ForestGreen!38}\textbf{\textcolor{black!85}{83.1}} & \cellcolor{ForestGreen!37}82.2 \\
  Vietnamese & 81.9 & \cellcolor{ForestGreen!11}83.2 & \cellcolor{ForestGreen!12}\textbf{\textcolor{black!85}{83.4}} & \cellcolor{ForestGreen!6}82.6 & 6.3 & \cellcolor{ForestGreen!36}78.6 & \cellcolor{ForestGreen!37}\textbf{\textcolor{black!85}{79.8}} & \cellcolor{ForestGreen!37}79.7 \\
  \textbf{\textcolor{black!85}{Average}} & 81.3 & \cellcolor{ForestGreen!5}82.0 & \cellcolor{ForestGreen!11}\textbf{\textcolor{black!85}{82.7}} & \cellcolor{ForestGreen!4}81.8 & 6.0 & \cellcolor{ForestGreen!35}77.0 & \cellcolor{ForestGreen!37}\textbf{\textcolor{black!85}{79.9}} & \cellcolor{ForestGreen!37}79.1 \\
  \midrule
  \multicolumn{9}{l}{\textbf{MLQA}} \\
  \cmidrule(lr){1-9}
  Arabic & 83.1 & \cellcolor{ForestGreen!8}84.1 & \cellcolor{ForestGreen!11}\textbf{\textcolor{black!85}{84.4}} & \cellcolor{ForestGreen!8}84.1 & 4.3 & \cellcolor{ForestGreen!36}76.3 & \cellcolor{ForestGreen!38}\textbf{\textcolor{black!85}{79.7}} & \cellcolor{ForestGreen!37}78.8 \\
  Chinese & 80.0 & \cellcolor{ForestGreen!7}80.9 & \cellcolor{ForestGreen!10}\textbf{\textcolor{black!85}{81.2}} & \cellcolor{ForestGreen!10}81.2 & 7.5 & \cellcolor{ForestGreen!27}62.4 & \cellcolor{ForestGreen!32}\textbf{\textcolor{black!85}{70.8}} & \cellcolor{ForestGreen!31}69.9 \\
  German & 80.9 & \cellcolor{ForestGreen!22}83.6 & \cellcolor{ForestGreen!24}\textbf{\textcolor{black!85}{83.9}} & \cellcolor{ForestGreen!22}83.6 & 11.8 & \cellcolor{ForestGreen!35}82.3 & \cellcolor{ForestGreen!36}\textbf{\textcolor{black!85}{84.5}} & \cellcolor{ForestGreen!36}83.3 \\
  Hindi & 81.0 & \cellcolor{ForestGreen!6}\textbf{\textcolor{black!85}{81.7}} & \cellcolor{ForestGreen!6}81.7 & \cellcolor{ForestGreen!3}81.4 & 12.7 & \cellcolor{ForestGreen!34}81.5 & \cellcolor{ForestGreen!35}\textbf{\textcolor{black!85}{83.1}} & \cellcolor{ForestGreen!35}82.4 \\
  Spanish & 82.8 & \cellcolor{ForestGreen!12}84.3 & \cellcolor{ForestGreen!13}\textbf{\textcolor{black!85}{84.5}} & \cellcolor{ForestGreen!12}84.3 & 10.7 & \cellcolor{ForestGreen!38}86.7 & \cellcolor{ForestGreen!39}\textbf{\textcolor{black!85}{88.6}} & \cellcolor{ForestGreen!39}88.3 \\
  Vietnamese & 82.1 & \cellcolor{ForestGreen!17}84.3 & \cellcolor{ForestGreen!19}\textbf{\textcolor{black!85}{84.4}} & \cellcolor{ForestGreen!17}84.2 & 13.9 & \cellcolor{ForestGreen!34}81.7 & \cellcolor{ForestGreen!35}\textbf{\textcolor{black!85}{83.7}} & \cellcolor{ForestGreen!35}83.2 \\
  \textbf{\textcolor{black!85}{Average}} & 81.6 & \cellcolor{ForestGreen!12}83.2 & \cellcolor{ForestGreen!14}\textbf{\textcolor{black!85}{83.3}} & \cellcolor{ForestGreen!12}83.1 & 10.1 & \cellcolor{ForestGreen!34}78.5 & \cellcolor{ForestGreen!36}\textbf{\textcolor{black!85}{81.7}} & \cellcolor{ForestGreen!35}81.0 \\
  \bottomrule
\end{tabular}
\caption{Ablations on the set of languages used for training, using our direct label projection evaluation schema in \S \ref{sec:label}. XML is used as the marker, with both the EN→XX and XX→EN directions evaluated and the results averaged. Base refers to the original unmodified model. We compare with models trained on three language sets: 1) one high resource language (One), 2) three high resource languages (Some), and 3) all seven languages (All). Differences with Base are highlighted in color.}
\label{tab:ablation-langs-label}
\end{table*}

The full Salesforce Localization XML MT dataset contains 7 languages with sentences parallel to English: German, Finnish, French, Japanese, Dutch, Russian, and Chinese. We conduct some basic ablations, training on translations both from and to English in the following combinations: 1) one high resource language (German), 2) three high-resource languages (German, Russian, Chinese), and 3) all seven languages. We evaluate label projection and translation quality in accordance with our methodology in \S \ref{sec:label}. 

The results are compiled in Table \ref{tab:ablation-langs-label}. We note that the model trained on three languages outperforms both the model trained on only one language and the model trained on all seven languages, in translation quality as well as label matches. We hypothesize that while the additional data helps improve the performance in the three-language model, including all seven languages induces catastrophic forgetting, thus reducing general performance. Given these results, we opt for the three-language model in all other experiments.

\section{Label Projection}
\label{sec:appendix:label}

\begin{table}[ht]
\centering
\begin{tabular}{lcc}
\toprule
 & \textbf{XQuAD} & \textbf{MLQA} \\
\midrule
\textbf{Samples (N)} & 2539 & 5414 \\
\textbf{Total Tags} & 23764 & 12265 \\
\textbf{Min Tags / Example} & 2 & 2 \\
\textbf{Max Tags / Example} & 24 & 8 \\
\textbf{Avg.\ \# Tags / Example} & 9.36 & 2.27 \\
\bottomrule
\end{tabular}
\caption{Tag statistics for evaluation datasets.}
\label{tab:eval-tag-stats}
\end{table}

Dataset statistics for the direct label projection evaluation datasets are given in Table \ref{tab:eval-tag-stats}. For Awesome-align, we use the label projection algorithm detailed by \citet{ebingDevilWordAlignment2025a} and BERT\textsubscript{base} \citep{devlinBERTPretrainingDeep2019} for word alignment itself. For Gemma 3 27B IT, we use XML tags as the markers for annotations and follow the setup of \citet{dabreStudyEffectivenessLarge2023}, using the prompt format given below, where \texttt{src\_lang} is the source language, \texttt{tgt\_lang} is the target language, and \texttt{src\_text} is the sentence to be translated:

\begin{center}
\fbox{\begin{minipage}{0.95\linewidth}
\small\sloppy
Translate the following \texttt{\{src\_lang\}} source text to \texttt{\{tgt\_lang\}}:\texttt{\textbackslash n}\\
\texttt{\{src\_lang\}}: \texttt{\{src\_text\}}\texttt{\textbackslash n}\\\texttt{\{tgt\_lang\}}:
\end{minipage}}
\end{center}

\subsection{MLQA Filtering}
\label{sec:appendix:mlqa}

\begin{table}[h]
  \centering
  \small
  \begin{tabular}{l r r r r r r}
    \toprule
    & \texttt{ar} & \texttt{de} & \texttt{es} & \texttt{hi} & \texttt{vi} & \texttt{zh} \\
    \midrule
    \textbf{Length} & 843 & 395 & 1152 & 908 & 1325 & 791 \\
    \bottomrule
  \end{tabular}
  \caption{Data statistics after filtering in MLQA.}
  \label{tab:mlqa-filtered-lengths}
\end{table}

Due to its nature as a dataset chiefly mined from Wikipedia, MLQA requires some filtering to act as a parallel label-projection evaluation benchmark. While questions and the sentences containing answers are aligned between languages, the paragraphs themselves are not necessarily direct translations. In order to make sure we only include paragraphs that are rough translations, we keep only paragraph pairs with the same number of questions and answer spans in both languages, and we filter out paragraphs with a COMET-22 score \citep{reiCOMET22UnbabelIST20222022} $<80$. The resulting dataset statistics are compiled in Table \ref{tab:mlqa-filtered-lengths}. We note that for the downstream evaluation in \S \ref{sec:downstream}, we use the full MLQA dataset as the filtering is not necessary for question-answering evaluation. 

\subsection{Label Projection into English}
\label{sec:appendix:label_xx_to_en}

\begin{table*}[t]
  \centering
  \small
  \begin{tabular}{l r r r r r r r r}
  \toprule
  \multirow[c]{2}{*}{\textbf{Language}} & \multicolumn{4}{c}{\textbf{COMET Score}} & \multicolumn{4}{c}{\textbf{Label Matches (F1, \%)}} \\
  \cmidrule(lr){2-5} \cmidrule(lr){6-9}
   & \textbf{Awes.} & \textbf{Gemma} & \textbf{EProj.} & \textbf{Ours} & \textbf{Awes.} & \textbf{Gemma} & \textbf{EProj.} & \textbf{Ours} \\
  \midrule
  \multicolumn{9}{l}{\textbf{XQUAD}} \\
  \cmidrule(lr){1-9}
  Arabic & 79.1 & \cellcolor{ForestGreen!10}\textbf{\textcolor{black!85}{81.6}} & \cellcolor{BrickRed!6}77.8 & \cellcolor{ForestGreen!6}80.6 & 36.0 & \cellcolor{ForestGreen!33}69.4 & \cellcolor{ForestGreen!24}59.9 & \cellcolor{ForestGreen!40}\textbf{\textcolor{black!85}{76.0}} \\
  Chinese & 80.3 & \cellcolor{ForestGreen!6}81.9 & \cellcolor{BrickRed!6}78.7 & \cellcolor{ForestGreen!10}\textbf{\textcolor{black!85}{82.7}} & 45.0 & \cellcolor{ForestGreen!20}64.7 & \cellcolor{ForestGreen!22}67.2 & \cellcolor{ForestGreen!35}\textbf{\textcolor{black!85}{80.0}} \\
  German & 82.3 & \cellcolor{ForestGreen!12}\textbf{\textcolor{black!85}{85.2}} & \cellcolor{BrickRed!4}81.3 & \cellcolor{ForestGreen!6}83.8 & 59.8 & \cellcolor{ForestGreen!20}80.3 & \cellcolor{ForestGreen!21}81.0 & \cellcolor{ForestGreen!26}\textbf{\textcolor{black!85}{85.6}} \\
  Greek & 80.9 & \cellcolor{ForestGreen!16}\textbf{\textcolor{black!85}{84.9}} & \cellcolor{BrickRed!3}80.3 & \cellcolor{ForestGreen!3}81.7 & 51.9 & \cellcolor{ForestGreen!22}73.4 & \cellcolor{ForestGreen!21}73.2 & \cellcolor{ForestGreen!25}\textbf{\textcolor{black!85}{77.3}} \\
  Hindi & 84.4 & \cellcolor{ForestGreen!1}84.8 & \cellcolor{BrickRed!4}83.3 & \cellcolor{ForestGreen!4}\textbf{\textcolor{black!85}{85.3}} & 52.0 & \cellcolor{ForestGreen!22}74.2 & \cellcolor{ForestGreen!26}77.9 & \cellcolor{ForestGreen!32}\textbf{\textcolor{black!85}{83.9}} \\
  Romanian & 81.1 & \cellcolor{ForestGreen!18}\textbf{\textcolor{black!85}{85.7}} & \cellcolor{ForestGreen!0}81.1 & \cellcolor{ForestGreen!8}83.0 & 56.5 & \cellcolor{ForestGreen!29}\textbf{\textcolor{black!85}{85.6}} & \cellcolor{ForestGreen!22}78.6 & \cellcolor{ForestGreen!26}82.6 \\
  Russian & 80.2 & \cellcolor{ForestGreen!11}\textbf{\textcolor{black!85}{83.1}} & \cellcolor{BrickRed!4}79.1 & \cellcolor{ForestGreen!6}81.6 & 50.4 & \cellcolor{ForestGreen!22}72.0 & \cellcolor{ForestGreen!25}75.1 & \cellcolor{ForestGreen!30}\textbf{\textcolor{black!85}{80.7}} \\
  Spanish & 83.3 & \cellcolor{ForestGreen!8}\textbf{\textcolor{black!85}{85.3}} & \cellcolor{BrickRed!2}82.7 & \cellcolor{ForestGreen!7}85.1 & 58.0 & \cellcolor{ForestGreen!21}79.3 & \cellcolor{ForestGreen!25}83.1 & \cellcolor{ForestGreen!29}\textbf{\textcolor{black!85}{87.1}} \\
  Thai & 81.5 & \cellcolor{ForestGreen!4}\textbf{\textcolor{black!85}{82.6}} & \cellcolor{BrickRed!10}79.2 & \cellcolor{BrickRed!5}80.3 & 34.5 & \cellcolor{ForestGreen!30}64.6 & \cellcolor{ForestGreen!25}59.7 & \cellcolor{ForestGreen!36}\textbf{\textcolor{black!85}{70.9}} \\
  Turkish & 82.3 & \cellcolor{ForestGreen!14}\textbf{\textcolor{black!85}{85.9}} & \cellcolor{BrickRed!1}82.1 & \cellcolor{ForestGreen!10}84.8 & 50.7 & \cellcolor{ForestGreen!31}82.0 & \cellcolor{ForestGreen!24}75.0 & \cellcolor{ForestGreen!32}\textbf{\textcolor{black!85}{82.9}} \\
  Vietnamese & 80.2 & \cellcolor{ForestGreen!17}\textbf{\textcolor{black!85}{84.5}} & \cellcolor{ForestGreen!0}80.2 & \cellcolor{ForestGreen!13}83.5 & 46.2 & \cellcolor{ForestGreen!36}\textbf{\textcolor{black!85}{82.2}} & \cellcolor{ForestGreen!29}74.8 & \cellcolor{ForestGreen!34}79.9 \\
  \textbf{\textcolor{black!85}{Average}} & 81.4 & \cellcolor{ForestGreen!11}\textbf{\textcolor{black!85}{84.1}} & \cellcolor{BrickRed!4}80.5 & \cellcolor{ForestGreen!6}82.9 & 49.2 & \cellcolor{ForestGreen!26}75.3 & \cellcolor{ForestGreen!24}73.2 & \cellcolor{ForestGreen!31}\textbf{\textcolor{black!85}{80.6}} \\
  \midrule
  \multicolumn{9}{l}{\textbf{MLQA}} \\
  \cmidrule(lr){1-9}
  Arabic & 81.9 & \cellcolor{BrickRed!13}78.7 & \cellcolor{BrickRed!4}81.0 & \cellcolor{ForestGreen!8}\textbf{\textcolor{black!85}{84.0}} & 39.6 & \cellcolor{ForestGreen!18}57.3 & \cellcolor{ForestGreen!24}63.6 & \cellcolor{ForestGreen!42}\textbf{\textcolor{black!85}{81.4}} \\
  Chinese & 80.3 & \cellcolor{BrickRed!8}78.3 & \cellcolor{BrickRed!2}79.7 & \cellcolor{ForestGreen!10}\textbf{\textcolor{black!85}{82.8}} & 37.6 & \cellcolor{ForestGreen!11}48.7 & \cellcolor{ForestGreen!24}62.1 & \cellcolor{ForestGreen!36}\textbf{\textcolor{black!85}{73.8}} \\
  German & 81.4 & \cellcolor{BrickRed!15}77.6 & \cellcolor{BrickRed!2}80.9 & \cellcolor{ForestGreen!9}\textbf{\textcolor{black!85}{83.7}} & 55.2 & \cellcolor{ForestGreen!11}66.2 & \cellcolor{ForestGreen!19}74.3 & \cellcolor{ForestGreen!30}\textbf{\textcolor{black!85}{85.7}} \\
  Hindi & 85.1 & \cellcolor{BrickRed!19}80.3 & \cellcolor{BrickRed!3}84.3 & \cellcolor{ForestGreen!5}\textbf{\textcolor{black!85}{86.4}} & 54.5 & \cellcolor{ForestGreen!4}58.2 & \cellcolor{ForestGreen!22}76.9 & \cellcolor{ForestGreen!32}\textbf{\textcolor{black!85}{86.8}} \\
  Spanish & 83.3 & \cellcolor{BrickRed!14}79.7 & \cellcolor{BrickRed!2}82.8 & \cellcolor{ForestGreen!6}\textbf{\textcolor{black!85}{84.9}} & 51.9 & \cellcolor{ForestGreen!8}59.4 & \cellcolor{ForestGreen!28}79.9 & \cellcolor{ForestGreen!36}\textbf{\textcolor{black!85}{88.3}} \\
  Vietnamese & 80.9 & \cellcolor{ForestGreen!0}81.0 & \cellcolor{ForestGreen!2}81.3 & \cellcolor{ForestGreen!13}\textbf{\textcolor{black!85}{84.2}} & 46.2 & \cellcolor{ForestGreen!18}63.7 & \cellcolor{ForestGreen!29}74.9 & \cellcolor{ForestGreen!39}\textbf{\textcolor{black!85}{85.0}} \\
  \textbf{\textcolor{black!85}{Average}} & 82.2 & \cellcolor{BrickRed!12}79.3 & \cellcolor{BrickRed!2}81.7 & \cellcolor{ForestGreen!9}\textbf{\textcolor{black!85}{84.3}} & 47.5 & \cellcolor{ForestGreen!11}58.9 & \cellcolor{ForestGreen!24}71.9 & \cellcolor{ForestGreen!36}\textbf{\textcolor{black!85}{83.5}} \\
  \bottomrule
\end{tabular}
\caption{Additional direct label projection results on XQuAD and MLQA, with sentences translated from the corresponding language to English. We compare four label projection methods: a) Awesome-align (Awes.), b) Gemma 3 27B (Gemma), c) EasyProject (EProj.), and d) LabelPigeon (LP). Awesome-align is used as the baseline, and differences are highlighted via color.}
\label{tab:direct_label_xxtoen}
\end{table*}

As label projection is largely applied for translating labeled data for low-resource languages, we focus on experiments with English as the source language. However, we also conduct a direct label projection experiment with English as the target language, compiled in Table \ref{tab:direct_label_xxtoen}. Here, LabelPigeon almost universally outperforms all other baselines in label matches, with EasyProject falling behind Gemma 3. Unlike our results in \S \ref{subsec:label:results}, Gemma 3 also provides strong translations with an average COMET score of $81.7$, outperforming EasyProject with $81.1$ and approaching LabelPigeon at $83.6$.

\subsection{Preliminary Baseline with \textsc{Codec}}
\label{sec:appendix:codec}
We did a comparison with \textsc{Codec} \citep{leConstrainedDecodingCrosslingual2024} in a preliminary experiment on the English-Hindi subset of XQuAD. \textsc{Codec} performed with a label match F1 of $75.6$, outperformed by LabelPigeon’s $76.9$. In addition, evaluation with \textsc{Codec} took significantly and prohibitively longer than any other tested method, roughly 38 minutes per sample. Given these results, we opted not to conduct a full-scale evaluation. In general, replicating the other label projection systems mentioned in \S \ref{sec:related_work} is challenging from an implementation standpoint, and their application is computationally expensive.

\section{Translation Quality}

In \S \ref{sec:translation}, we synthetically insert markers into the \textsc{Flores-200} dataset to test the impact of our method on translation quality. Expanded results and additional details are provided below.

\subsection{Synthetic Marker Insertion}
\label{sec:appendix:synthetic_marker}
We model this process by iterating through the word boundaries in the sentence. At each word boundary, an open marker may be placed with a probability of $P_{open}$, starting a new label span. If a label span has already been started, at each subsequent word boundary a close marker may be placed with a probability of $P_{close}$, ending the span. If any spans are open by the end of the sentence, the appropriate close markers are inserted at the end. We refer to this as the \textbf{Complex} marker insertion configuration, as nesting and overlapping spans are possible. By preventing new spans from being started if a span is already open, we disable nesting and overlapping, and we refer to this as the \textbf{Simple} configuration. To simulate datasets with exactly one labeled span per sample, we first sample a length \(L \sim \mathrm{Geom}(P_{close})\), and then select a span uniformly at random among all candidate spans of length \(L\) in the sentence. We refer to this as the \textbf{Single} configuration. In general, the $P_{open}$ and $P_{close}$ allow us to model the frequency of labels and their average length, respectively. These values are set to $0.2$ and $0.5$ for all our experiments unless specified otherwise. 

\subsection{Full \textsc{Flores-200} Results}
\label{sec:appendix:full_flores_results}

We provide the full results of our \textsc{Flores-200} experiments in Tables \ref{tab:full_flores_0_100} and \ref{tab:full_flores_101_204}. We note that the performance improvement of the fine-tuned models are largely consistent across all languages, the vast majority of which are unseen during fine-tuning.

\subsection{Variation with Marker Frequency and Length}
\label{sec:appendix:marker_freq}

\begin{figure}[t]
  \includegraphics[width=\columnwidth]{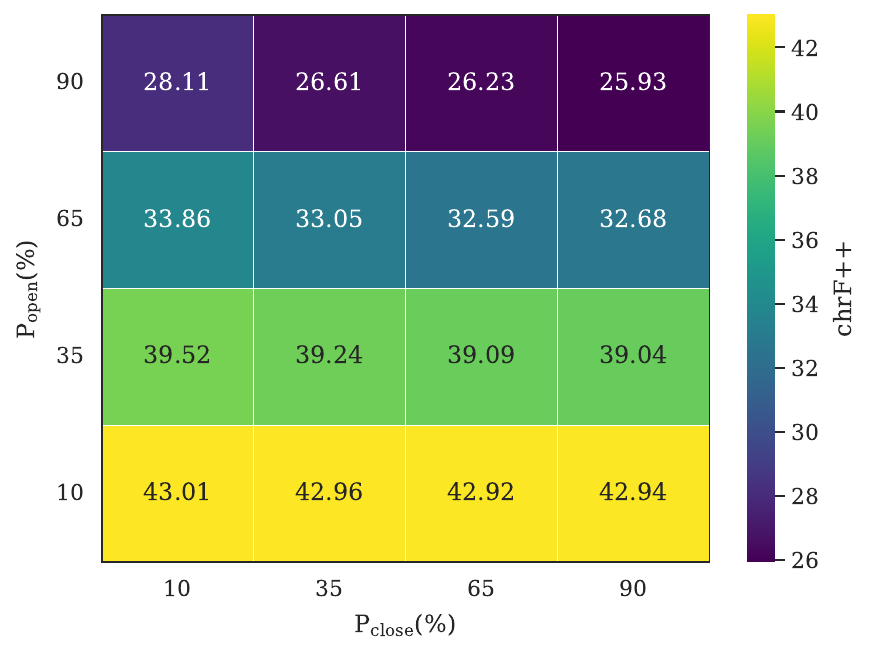}
  \caption{Translation performance of our model on \textsc{Flores-200} as measured by chrF++ across different values of $P_{close}$ and $P_{open}$ under the Complex marker insertion scheme.}
  \label{fig:flores_hyperparam_heatmap}
\end{figure}

We also estimate the effect of the frequency and length of the marked spans on translation quality by varying $P_{open}$ and $P_{close}$, specifically in the Complex marker insertion configuration. Figure \ref{fig:flores_hyperparam_heatmap} compiles the results as a heatmap. We see a clear degradation of quality with increasing tag frequency, but a slight and consistent improvement with increasing length (i.e., with decreasing $P_{close}$). Nevertheless, we note that at the lowest tag frequency (corresponding to roughly one tag once every ten words), translation quality is still improved from the baseline.

\section{Downstream Experiments}
\label{sec:appendix:downstream}

For downstream experiments, we utilize already available baselines and corresponding code, making minimal changes. For NER, we use the scripts provided by \citet{chenFrustratinglyEasyLabel2024}, training 5 epochs with a batch size of 32 and a learning rate of 2e\textsuperscript{-3}. We average the result of five random seeds to minimize variance.

For QA, we use the scripts provided by \citet{huXTREMEMassivelyMultilingual2020}, training 5 epochs with a batch size of 32 and learning rate of 3e\textsuperscript{-3}, over three random seeds. We also discard samples from SQuAD that have more than one missing question-answer span after translation to ensure high data quality. 

For coreference resolution, we use the scripts provided for the CRAC shared task \citep{novakFindingsThirdShared2024}, training 5 epochs with a batch size of 1 document and a learning rate of 2e\textsuperscript{-4} for task-specific parameters and 1e\textsuperscript{-5} for others. Additionally, to speed up the evaluation, we conduct a simple filtering step on OntoNotes, retaining documents with six sentences or fewer, in line with the default maximum sentence limit that the downstream model handles. We also note that the metric is specifically exact-match F1 excluding singletons.

\section{License}
We use several datasets under various licenses in this work, which we enumerate below.

\begin{itemize}

    \item \textbf{XQuAD}: \href{https://creativecommons.org/licenses/by-sa/4.0/}{CC BY-SA 4.0}

    \item \textbf{MLQA}: \href{https://creativecommons.org/licenses/by-sa/3.0/}{CC BY-SA 3.0}
    
    \item \textbf{\textsc{Flores-200}}: \href{https://creativecommons.org/licenses/by-sa/4.0/}{CC BY-SA 4.0}

    \item \textbf{UNER}: \href{https://creativecommons.org/licenses/by-sa/4.0/}{CC BY-SA 4.0}

    \item \textbf{CorefUD}: \href{https://lindat.mff.cuni.cz/repository/static/license-corefud-1.3.html}{License CorefUD v1.3}

    \item \textbf{SQuAD}: \href{https://creativecommons.org/licenses/by-sa/4.0/}{CC BY-SA 4.0}

    \item \textbf{OntoNotes}: \href{https://catalog.ldc.upenn.edu/license/ldc-non-members-agreement.pdf}{LDC User Agreement for Non-Members}

\end{itemize}

All datasets used were employed in accordance with their intended research purposes and license terms. All created artifacts are intended for research and academic dissemination consistent with these terms.

\begin{table*}[t]
\centering
\tiny
\resizebox{\textwidth}{!}{%
\begin{tabular}{lrrrrrrrrrrrrrrrrrrrr}
\toprule
 & \multicolumn{10}{c}{\textbf{BLEU}} & \multicolumn{10}{c}{\textbf{chrF++}} \\
\cmidrule(lr){2-11} \cmidrule(lr){12-21}
\textbf{Language} & \multicolumn{4}{c}{\textbf{No Markers}} & \multicolumn{2}{c}{\textbf{Single}} & \multicolumn{2}{c}{\textbf{Simple}} & \multicolumn{2}{c}{\textbf{Complex}} & \multicolumn{4}{c}{\textbf{No Markers}} & \multicolumn{2}{c}{\textbf{Single}} & \multicolumn{2}{c}{\textbf{Simple}} & \multicolumn{2}{c}{\textbf{Complex}} \\
\cmidrule(lr){2-5} \cmidrule(lr){6-7} \cmidrule(lr){8-9} \cmidrule(lr){10-11} \cmidrule(lr){12-15} \cmidrule(lr){16-17} \cmidrule(lr){18-19} \cmidrule(lr){20-21}
 & \textbf{Baseline} & \textbf{EProj.} & \textbf{LP} & \textbf{NF} & \textbf{EProj.} & \textbf{LP} & \textbf{EProj.} & \textbf{LP} & \textbf{EProj.} & \textbf{LP} & \textbf{Baseline} & \textbf{EProj.} & \textbf{LP} & \textbf{NF} & \textbf{EProj.} & \textbf{LP} & \textbf{EProj.} & \textbf{LP} & \textbf{EProj.} & \textbf{LP} \\
\midrule
\texttt{ace\_Arab} & 0.8 & \cellcolor{BrickRed!3}0.7 & \cellcolor{BrickRed!1}0.8 & \cellcolor{BrickRed!3}0.7 & \cellcolor{BrickRed!4}0.6 & \cellcolor{BrickRed!4}0.6 & \cellcolor{BrickRed!7}0.4 & \cellcolor{BrickRed!6}0.5 & \cellcolor{BrickRed!9}0.3 & \cellcolor{BrickRed!6}0.5 & 18.1 & \cellcolor{ForestGreen!1}18.2 & \cellcolor{BrickRed!8}17.0 & \cellcolor{BrickRed!7}17.2 & \cellcolor{ForestGreen!0}18.1 & \cellcolor{BrickRed!15}16.2 & \cellcolor{BrickRed!6}17.3 & \cellcolor{BrickRed!9}16.9 & \cellcolor{BrickRed!7}17.2 & \cellcolor{BrickRed!8}17.1 \\
\bottomrule
\end{tabular}
}
\caption{Full results across different marker insertion configurations on the \textsc{Flores-200} dataset (Languages 0–99).}
\label{tab:full_flores_0_100}
\end{table*}

\begin{table*}[t]
\centering
\tiny
\resizebox{\textwidth}{!}{%
\begin{tabular}{lrrrrrrrrrrrrrrrrrrrr}
\toprule
 & \multicolumn{10}{c}{\textbf{BLEU}} & \multicolumn{10}{c}{\textbf{chrF++}} \\
\cmidrule(lr){2-11} \cmidrule(lr){12-21}
\textbf{Language} & \multicolumn{4}{c}{\textbf{No Markers}} & \multicolumn{2}{c}{\textbf{Single}} & \multicolumn{2}{c}{\textbf{Simple}} & \multicolumn{2}{c}{\textbf{Complex}} & \multicolumn{4}{c}{\textbf{No Markers}} & \multicolumn{2}{c}{\textbf{Single}} & \multicolumn{2}{c}{\textbf{Simple}} & \multicolumn{2}{c}{\textbf{Complex}} \\
\cmidrule(lr){2-5} \cmidrule(lr){6-7} \cmidrule(lr){8-9} \cmidrule(lr){10-11} \cmidrule(lr){12-15} \cmidrule(lr){16-17} \cmidrule(lr){18-19} \cmidrule(lr){20-21}
 & \textbf{Baseline} & \textbf{EProj.} & \textbf{LP} & \textbf{NF} & \textbf{EProj.} & \textbf{LP} & \textbf{EProj.} & \textbf{LP} & \textbf{EProj.} & \textbf{LP} & \textbf{Baseline} & \textbf{EProj.} & \textbf{LP} & \textbf{NF} & \textbf{EProj.} & \textbf{LP} & \textbf{EProj.} & \textbf{LP} & \textbf{EProj.} & \textbf{LP} \\
\midrule
\texttt{kor\_Hang} & 11.6 & \cellcolor{ForestGreen!18}12.7 & \cellcolor{BrickRed!18}10.4 & \cellcolor{BrickRed!12}10.8 & \cellcolor{BrickRed!15}10.6 & \cellcolor{ForestGreen!9}12.1 & \cellcolor{BrickRed!33}9.5 & \cellcolor{BrickRed!24}10.1 & \cellcolor{BrickRed!37}9.3 & \cellcolor{BrickRed!34}9.4 & 33.1 & \cellcolor{ForestGreen!7}34.0 & \cellcolor{BrickRed!4}32.7 & \cellcolor{BrickRed!0}33.1 & \cellcolor{BrickRed!6}32.3 & \cellcolor{BrickRed!4}32.6 & \cellcolor{BrickRed!27}29.7 & \cellcolor{BrickRed!26}29.9 & \cellcolor{BrickRed!29}29.5 & \cellcolor{BrickRed!33}29.0 \\
\bottomrule
\end{tabular}
}
\caption{Contd. results across different marker insertion configurations on the \textsc{Flores-200} dataset (languages 100–203).}
\label{tab:full_flores_101_204}
\end{table*}


\end{document}